\documentclass[preprint,7pt,times,twocolumn]{article}

\usepackage[a4paper, total={7.5in, 7.5in}]{geometry}

\usepackage{mathrsfs}

\usepackage{subcaption}

\usepackage{color}
\usepackage{graphicx,hyperref}
\usepackage{amsmath}
\usepackage{amssymb}

\newcommand{\real}{\mbox{\rm I$\!$R}}

\newcommand{\bld}[1]{\mbox{\boldmath $#1$}} 

\newcommand{\x}{\bld{x}}
\newcommand{\F}{\bld{F}}

\newcommand{\M}{\bld{M}}

\usepackage[ruled,vlined,lined,linesnumbered,english]{algorithm2e}

\SetCommentSty{mycommfont}
\SetKwProg{Fn}{Function}{}{}

\title{Analytical Forward Dynamics Modeling of Linearly Actuated Heavy-Duty Parallel-Serial Manipulators\footnote{This work was supported by The Business Finland partnership project “Future all-electric rough terrain autonomous mobile manipulators” (Grant 2334/31/2022). \\ \textit{alvaro.pazanaya@tuni.fi, jouni.mattila@tuni.fi}}}

\author{Alvaro Paz, Jouni Mattila \\ {\scriptsize Faculty of Engineering and Natural Sciences, Tampere University. Tampere, Finland.}}

\begin{document}

\date{}
\maketitle

\begin{abstract}
This paper presents a new geometric and recursive algorithm for analytically computing the forward dynamics of heavy-duty parallel-serial mechanisms. Our solution relies on expressing the dynamics of a class of linearly-actuated parallel mechanism to a lower dimensional dual Lie algebra to find an analytical solution for the inverse dynamics problem. Thus, by applying the articulated-body inertias method, we successfully provide analytic expressions for the total wrench in the linear-actuator reference frame, the linear acceleration of the actuator, and the total wrench exerted in the base reference frame of the closed loop. This new formulation allows to backwardly project and assemble inertia matrices and wrench bias of multiple closed-loops mechanisms. The final algorithm holds an $\mathcal{O}(n)$ algorithmic complexity, where $n$ is the number of degrees of freedom (DoF). We provide accuracy results to demonstrate its efficiency with 1-DoF closed-loop mechanism and 4-DoF manipulator composed by serial and parallel mechanisms. Additionally, we release a URDF multi-DoF code for this recursive algorithm.\footnote{The open-source implementation of the algorithm in Matlab Simscape will be provided at github.com/AlvaroPaz/SerialParallelForwardDynamics.}
\end{abstract}



%
%
%
%

%
\section{Introduction}

Heavy-duty machinery modeling and control are of great importance due to their widely spread industrial high-payloads applications \cite{abdel2003dynamics}\cite{mattila2017survey} (e.g., foresting, constructing, and mining). Among such mechanical systems, the parallel-serial manipulators are a class of multiple degrees-of-freedom (DoF) robots powered by linear actuators, hydraulic or electric, whose kinematic topology is composed by a serial concatenation of parallel mechanisms, closed loop kinematic chain, and serial mechanisms (e.g., prismatic and rotational motion).

Moreover, motion generation of these widely used and important parallel-serial manipulators, and their closed-loop kinematic chains, has been investigated through model-based control frameworks \cite{lewis2003robot}\cite{zhu2010virtual}\cite{Bib:murray}. In particular, model predictive control (MPC) approaches, which implement sophisticated and subtle techniques, require by their nature not only compute the robot's dynamics in an accurate manner but also in a fast way for achieving real-time decisions making \cite{diehl2009efficient}\cite{diehl2006fast}. Such accuracy and speed can be accomplished with tailored algorithms for computing the system's dynamics in analytical manner. Additionally, since nonlinear numerical-optimization based controls require to compute not only the system dynamics but also its first-and-second order partial derivatives, with respect to the decision variable. Thus, some optimization solvers based on automatic differentiation have demonstrated efficient results \cite{andersson2019casadi}, with the main drawback that they usually require a pre-computation stage each time the robot topology changes. Also, numeric analysis formulations, in particular Kane's method, have reported good performance for solving the dynamics of hydraulic excavators \cite{vsalinic2014dynamic}.

However, recursive algorithms represent a compact, elegant and flexible, formulation for solving robot's dynamics. On one hand, spatial six-dimensional (6D) vectors, which are based in Plücker coordinates, can completely represent both linear and angular motions of multibody mechanisms in a compact and recursive formulation while avoiding complex closed-forms computation (e.g., the generalized inertia and Coriolis matrices). For instance, concerning inverse and forward dynamics, the recursive Newton Euler algorithm (RNEA) and the articulated body algorithm (ABA) \cite{Bib:Featherstone} are efficient options in terms of algorithmic complexity, respectively. On the other hand, a geometric approach, based on screw theory \cite{Bib:selig}\cite{Bib:Park}, can completely formulate these recursive algorithms for computing the robot's kinematics and dynamics \cite{muller2018screwKinematics}\cite{muller2018screwDynamics} by means of Lie groups and their algebras with some benefits, such as direct partial differentiation, due to its exponential maps \cite{carpentier2018analytical}\cite{paz2023analytical}.

Additionally, the kinematics and dynamic reaction forces of four-links parallel mechanisms (see Figure \ref{fig:parallel}), composed by three rotational passive joints and one linear actuated joint, can be studied using screw theory as in \cite{cibicik2019dynamic}. Also, the inverse dynamics of these mechanisms is presented in \cite{zhu2010virtual} by following Newton-Euler formalism in a virtual decomposition procedure and imposing zero-torque constraints at passive joints. However, this solution relies on finding load distribution factors for internal forces in the closed kinematic chain. Therefore, a geometric approach for finding such factors in analytic form has been proposed \cite{koivumaki2018addressing}. For the same 4-joints parallel mechanism, more recent works \cite{petrovic2022mathematical} successfully presented an analytic solution for computing the linear actuator force by assuming new constraints in the mechanism wrenches that eliminate the internal wrenches dependency and thus the calculation of distribution factors.

By its nature, some trajectory optimization approaches (i.e., multiple shooting \cite{diehl2006fast}) require to compute the forward dynamics of the system rather than the inverse. This dynamics for tree-topology kinematic chains can be efficiently solved by means of ABA, which is mainly based in the articulated-body inertias method \cite{Bib:Featherstone}. Nevertheless, parallel kinematic chains usually require considering a set of holonomic constraints for loop closure, which for most of parallel mechanisms is solved numerically. This numerical solutions are not only inaccurate but also increase the computational burden. Efforts on finding completely analytic solutions for parallel-serial manipulators have been investigated; for instance, \cite{jain2010robot} presented a local-constraints embedding approach for general closed-loops mechanism by following a cutting-edges procedure, also \cite{kumar2022modular} reported a hybrid solution (numeric and analytic) for these manipulators where closure equations are solved analytically when possible.

The contributions of this paper are as follows. For this class of heavy-duty manipulators, which contain parallel mechanisms, like that shown in Figure \ref{fig:parallel}, we first take advantage of the solution in \cite{petrovic2022mathematical} to project all parallel mechanism dynamics in a lower dimensional dual Lie algebra $se^{*}(2)$ and generate an analytic solution for the total wrench exerted at linear actuator's reference frame. This allows finding a linear and scalar expression to solve the inverse dynamics of parallel mechanisms and to forward dynamics in analytic expression. This is achieved by implementing the articulated-body inertias method, and consequently our algorithm inherits the ABA benefits, such as linear arithmetic complexity $\mathcal{O}(n)$, while avoiding closed-forms computation and matrix inversions.

Additionally, by following the articulated-body inertias method, we formulate expressions for the total assembled inertias and wrench bias acting on the base reference frame of the parallel mechanism. Such expressions allow to concatenate multiple closed kinematics loops in a serial sequence and solve its forward dynamics. Thus, we provide an explicit recursive-and-exact algorithm for solving the forward dynamics of multi-DoF parallel-serial manipulators that are composed by a sequence of parallel mechanisms and simple 1-DoF serial mechanisms (i.e., prismatic and rotational motions). We also provide a URDF-based code repository for this recursive algorithm.

The paper is organized as follows. Section \ref{sec:dynamics_parallel} presents in a geometric way (screw theory) the kinematics and dynamics of parallel mechanisms and expressions for the total wrench in the actuator's reference frame. Assembly inertias method in Section \ref{sec:3} to find both a linear and scalar expression for the actuator's acceleration and for the assembled inertias expressions in the base reference frame of the closed loop. Section \ref{sec:manipulator} presents the complete algorithm for the parallel-serial manipulator whether it is composed by parallel or serial mechanisms. The results for testing the accuracy of our algorithm are reported in Section \ref{sec:results} for 1-DoF parallel mechanism and a 4-DoF parallel-serial manipulator. Some conclusions are pointed out in Section \ref{sec:conclusions}.

\section{Dynamics of parallel mechanisms}
\label{sec:dynamics_parallel}

\begin{figure}[h!] 
	\centering
	\includegraphics[width=8cm]{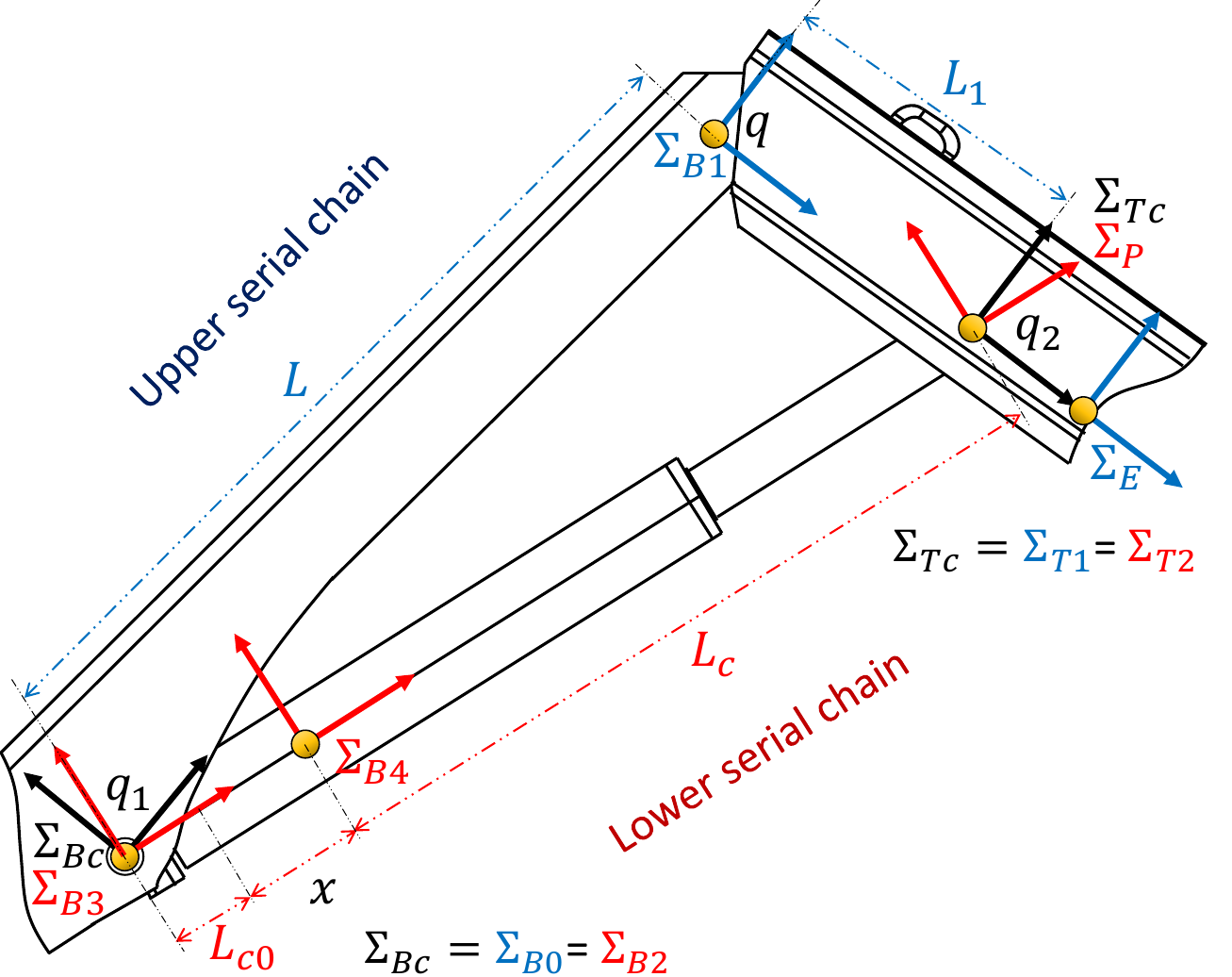}
	\vspace*{-0.2cm}
	\caption{{\footnotesize {\bf Parallel mechanism}. It is a 1-DoF closed chain composed by four joints, four links, and one linear actuator. We also attach eleven reference frames $\Sigma$ for our analysis, and in a kinematic sense, this mechanism can be considered as two serial kinematic chains, upper and lower.}}
	\label{fig:parallel}
\end{figure}

A linearly actuated parallel mechanism is defined as a closed kinematic chain composed by four joints (i.e., three passive revolute and one prismatic actuated). See Figure \ref{fig:parallel}.

Let's distinguish the following reference frames for our analysis. The upper serial kinematic chain is $\Sigma_{B0}$, $\Sigma_{B1}$, $\Sigma_{T1}$ and $\Sigma_{E}$, and the lower one is $\Sigma_{B2}$, $\Sigma_{B3}$, $\Sigma_{B4}$, $\Sigma_{P}$ and $\Sigma_{T2}$. At the beginning and ending of the closed kinematic chain, we can place general reference frames as
\begin{equation}
	\Sigma_{Bc} \ = \ \Sigma_{B0} \ = \ \Sigma_{B2}
	\qquad \mbox{and} \qquad \Sigma_{Tc} \ = \ \Sigma_{T1} \ = \ \Sigma_{T2}
	\label{eq:frames}
\end{equation}

The topology of this mechanism is thus defined by the connectivity vectors, acording to \cite{Bib:Featherstone} nomenclature
\begin{eqnarray*}
		i & = & [ B0, B1, T1, E, B2, B3, B4, P, T2 ] \\
		\lambda & = & [ -, B0, B1, T1, -, B2, B3, B4, P ] \\
		\mu & = & [ B1, T1, E, -, B3, B4, P, T2, E ]
\end{eqnarray*}
where $i$ is the reference frame index, $\lambda$ is its corresponding successor, and $\mu$ is its predecessor.

\subsection{Closed-chain kinematics}
The mechanism configuration is parameterized by the passive joint angles $q$, $q_1$, and $q_2$ and by the actuated joint position $x\geq0$. Such passive joints are placed at the origin of frames $\Sigma_{B1}$, $\Sigma_{B3}$, and $\Sigma_{T2}$ and the actuated one in $\Sigma_{B4}$.

The geometric closure constraints are forced through the next expressions in the configuration \cite{zhu2010virtual}
\begin{subequations}
	\begin{eqnarray}
		x & = & \sqrt{L^{2}\!+\!L_{1}^{2}\!+\!2LL_{1}\cos q} \!-\! x_{0} \\
		q_{1} & = & -\!\arccos\!\left(\! \tfrac{L_{1}^{2}-(x+x_{0})^{2}-L^{2}}{-2(x+x_{0})L} \right) \\
		q_{2} & = & -\!\arccos\!\left(\! \tfrac{L^{2}-(x+x_{0})^{2} - L_{1}^{2}}{-2(x+x_{0})L_{1}} \right)
	\end{eqnarray}
\end{subequations}
where $L, L_{1} > 0$ and $x_{0}=L_{c}+L_{c0}$ are link lengths. See Figure \ref{fig:parallel}.

The forward kinematics is given by the homogeneous transformation matrices $\bld{G}$, which are elements of the Lie group $SE(3)$. The associated Lie algebra of this group is $se(3)$, which contains motion elements denoted as $\bld{\xi}$, e.g. screw vector $\bld{s}$, twist $\bld{\nu}$ and spatial acceleration $\dot{\bld{\nu}}$. Also, its associated dual Lie algebra $se^{*}(3)$ contains force elements $\bld{\xi}^{*}$ (e.g., spatial momentum $\bld{\mu}$ and wrench vector $\bld{F}$).

Moreover, the transformation matrix $\bld{G}$ and the adjoint operator $\mbox{Ad}_{G}$ are representations of the elements in $SE(3)$ and can be computed by means of the exponential forms \cite{Bib:selig}
\begin{equation}
	\bld{G} \ = \ e^{[\bld{s}]q}
	\qquad \mbox{and} \qquad \mbox{Ad}_{e^{[{\scriptsize \bld{s}}]q}} \ = \ e^{\mbox{ad}_{{\scriptsize \bld{s}} q}}
	\label{eq:op00}
\end{equation}
where $q$ is the joint position, and $\mbox{ad}_{(\cdot)}$ is the Lie bracket operator. For more operators details see \cite{paz2023analytical}.

Joint velocities are constrained by the following first order expressions
\begin{equation}
	\dot{q} \ = \ k_1 \dot{x} \qquad\quad \dot{q}_{1} \ = \ k_2 \dot{x} \qquad\quad \dot{q}_{2} \ = \ k_3 \dot{x}
	\label{eq:dq__}
\end{equation}
where
\begin{subequations}
	\begin{eqnarray}
		k_1 & = & -\left( x+x_{0} \right) / \left(\!LL_{1}\!\sin q\right) \\
		k_2 & = & -\!\left(x\!+\!x_{0} \!-\! L\!\cos q_{1}\right)/(x\!+\!x_{0})L\sin q_{1} \\
		k_3 & = & -\!\left(x\!+\!x_{0} \!-\! L_{1}\!\cos q_{2}\right)/(x\!+\!x_{0})L_{1}\sin q_{2}
	\end{eqnarray}
	\label{eq:k1_3}
\end{subequations}
Note that $q, q_{1}$ and $ q_{2} \neq 0$ for singularity avoidance.

According to definition (\ref{eq:frames}), frames $\Sigma_{B0}$ and $\Sigma_{B2}$ have the same twist $\bld{\nu}_{\!_{\!Bc}}=\bld{\nu}_{\!_{\!B0}}=\bld{\nu}_{\!_{\!B2}}$. Then, twists in the most representative reference frames are computed as
\begin{subequations}
	\begin{eqnarray}
		\bld{\nu}_{\!_{\!B1}} & = & \mbox{Ad}_{G_{\!_{B1}}^{\!_{B0}}}\bld{\nu}_{\!_{\!B0}} + \bld{s}_{z}\dot{q} \label{eq:nu_b1} \\
		\bld{\nu}_{\!_{\!E}} & = & \mbox{Ad}_{G_{\!_{E}}^{\!_{B1}}}\bld{\nu}_{\!_{\!B1}} \label{eq:nu_e} \\
		\bld{\nu}_{\!_{\!B3}} & = & \mbox{Ad}_{G_{\!_{B3}}^{\!_{B2}}}\bld{\nu}_{\!_{\!B2}} + \bld{s}_{z}\dot{q}_{1} \label{eq:nu_b3} \\
		\bld{\nu}_{\!_{\!B4}} & = & \mbox{Ad}_{G_{\!_{B4}}^{\!_{B3}}}\bld{\nu}_{\!_{\!B3}} + \bld{s}_{x}\dot{x} \label{eq:nu_b4} \\
		\bld{\nu}_{\!_{\!T2}} & = & \mbox{Ad}_{G_{\!_{T2}}^{\!_{B4}}}\bld{\nu}_{\!_{\!B4}} + \bld{s}_{z}\dot{q}_{2} \label{eq:nu_t2}
	\end{eqnarray}
	\label{eq:36}
\end{subequations}
where the screw vectors are
\begin{eqnarray*}
	\bld{s}_{x} & = & \begin{bmatrix}
		1 & 0 & 0 & 0 & 0 & 0
	\end{bmatrix}^{\top} \\
	\bld{s}_{z} & = & \begin{bmatrix}
		0 & 0 & 0 & 0 & 0 & 1
	\end{bmatrix}^{\top}
\end{eqnarray*}

Joint acceleration constraints are found by deriving (\ref{eq:dq__}) as
\begin{equation}
	\ddot{q} = \dot{k}_1 \dot{x} + k_1 \ddot{x} \qquad \ddot{q}_{1} = \dot{k}_2 \dot{x} + k_2 \ddot{x} \qquad \ddot{q}_{2} = \dot{k}_3 \dot{x} + k_3 \ddot{x}
	\label{eq:04}
\end{equation}
where $\dot{k}_1$, $\dot{k}_2$, and $\dot{k}_3$ are the time derivatives of (\ref{eq:k1_3}), and $\ddot{x}$ is the actuator's linear acceleration.

Additionally from definition (\ref{eq:frames}), frames $\Sigma_{B0}$ and $\Sigma_{B2}$ have the same spatial acceleration
\begin{equation}
	\dot{\bld{\nu}}_{\!_{\!Bc}} \ = \ \dot{\bld{\nu}}_{\!_{\!B0}} \ = \ \dot{\bld{\nu}}_{\!_{\!B2}}
	\label{eq:07}
\end{equation}
Then, spatial accelerations are recursively computed as
\begin{subequations}
	\begin{eqnarray}
		\dot{\bld{\nu}}_{\!_{\!B1}} & = & \mbox{Ad}_{G_{\!_{B1}}^{\!_{B0}}}\dot{\bld{\nu}}_{\!_{\!B0}} + \bld{s}_{z}\ddot{q} + \bld{c}_{\!_{B1}} \label{eq:dnu_b1} \\
		\dot{\bld{\nu}}_{\!_{\!E}} & = & \mbox{Ad}_{G_{\!_{E}}^{\!_{B1}}}\dot{\bld{\nu}}_{\!_{\!B1}} \label{eq:dnu_e} \\
		\dot{\bld{\nu}}_{\!_{\!B3}} & = & \mbox{Ad}_{G_{\!_{B3}}^{\!_{B2}}}\dot{\bld{\nu}}_{\!_{\!B2}} + \bld{s}_{z}\ddot{q}_{1} + \bld{c}_{\!_{B3}} \label{eq:dnu_b3} \\
		\dot{\bld{\nu}}_{\!_{\!B4}} & = & \mbox{Ad}_{G_{\!_{B4}}^{\!_{B3}}}\dot{\bld{\nu}}_{\!_{\!B3}} + \bld{s}_{x}\ddot{x} + \bld{c}_{\!_{B4}} \label{eq:dnu_b4} \\
		\dot{\bld{\nu}}_{\!_{\!T2}} & = & \mbox{Ad}_{G_{\!_{T2}}^{\!_{B4}}}\dot{\bld{\nu}}_{\!_{\!B4}} + \bld{s}_{z}\ddot{q}_{2} + \bld{c}_{\!_{T2}} \label{eq:dnu_t2}
	\end{eqnarray}
	\label{eq:53}
\end{subequations}
where $\bld{c}_{i}$ is the spatial acceleration bias
\begin{equation}
	\bld{c}_{i} \ = \ \mbox{ad}_{\nu_i}\bld{s}_i\dot{q}_i
	\label{eq:37}
\end{equation}
\subsection{Closed-chain dynamics}
The reference frames $\Sigma_{B0}$, $\Sigma_{B1}$, $\Sigma_{B3}$, $\Sigma_{B4}$, and $\Sigma_{E}$ have attached inertial bodies. Thus the local wrenches denoted by $\hat{\F}_{\!_{\!B0}}$, $\hat{\F}_{\!_{\!B1}}$, $\hat{\F}_{\!_{\!B3}}$, $\hat{\F}_{\!_{\!B4}}$, and $\hat{\F}_{\!_{\!E}}$, respectively, can be computed by using the controlled Euler-Poincar\'{e} equation \cite{marsden2013introduction}
\begin{equation}
	\hat{\F}_i \ = \ \M_i\dot{\bld{\nu}}_i + \bld{p}_{i}
	\label{eq:wrench_i}
\end{equation}
where $\M_i \in \real^{6\times 6}$ is the spatial inertia matrix, $\mbox{ad}_{(\cdot)}^{*}$ is the adjoint dual operator, and $\bld{p}_{i}$ is the wrench bias
\begin{equation}
	\bld{p}_{i} \ = \ - \mbox{ad}_{\nu_i}^* \M_i\bld{\nu}_i
	\label{eq:38}
\end{equation}

By considering that $\F_{\!_{\!P}}$ is the total force in frame $\Sigma_{P}$ and using the adjoint dual operator $\mbox{Ad}_{G}^{*}$ for transforming wrench vectors between reference frames, the total upper-kinematic-chain wrenches are solved as \cite{petrovic2022mathematical}
\begin{subequations}
	\begin{eqnarray}
		\F_{\!_{\!B1}} & = & \hat{\F}_{\!_{\!B1}} + \mbox{Ad}_{G_{\!_{E}}^{\!_{B1}}}^{*} \F_{\!_{\!E}} - \mbox{Ad}_{G_{\!_{P}}^{\!_{B1}}}^{*} \F_{\!_{\!P}} \\
		\F_{\!_{\!B0}} & = & \hat{\F}_{\!_{\!B0}} + \mbox{Ad}_{G_{\!_{B1}}^{\!_{B0}}}^{*} \F_{\!_{\!B1}}
	\end{eqnarray}
	\label{eq:dyn_01}
\end{subequations}
and for the lower kinematic chain
\begin{subequations}
	\begin{eqnarray}
		\F_{\!_{\!B4}} & = & \hat{\F}_{\!_{\!B4}} + \mbox{Ad}_{G_{\!_{P}}^{\!_{B4}}}^{*} \F_{\!_{\!P}} \\
		\F_{\!_{\!B3}} & = & \hat{\F}_{\!_{\!B3}} + \mbox{Ad}_{G_{\!_{B4}}^{\!_{B3}}}^{*} \F_{\!_{\!B4}} \\
		\F_{\!_{\!B2}} & = & \mbox{Ad}_{G_{\!_{B3}}^{\!_{B2}}}^{*} \F_{\!_{\!B3}}
	\end{eqnarray}
	\label{eq:dyn_02}
\end{subequations}

Then the total force at $\Sigma_{Bc}$ is
\begin{equation}
	\F_{\!_{\!Bc}} \ = \ \F_{\!_{\!B0}} + \F_{\!_{\!B2}}
	\label{eq:dyn_03}
\end{equation}
and the linear actuator force $f_{\!_{B4}}^{x}$ is computed as
\begin{equation}
	f_{\!_{B4}}^{x} \ = \ \bld{s}_{x}^{\top} \F_{\!_{\!B4}}
	\label{eq:dyn_04}
\end{equation}

See \ref{app:inv_dyn} for an analytical solution of $f_{\!_{B4}}^{x}$.
\section{The assembly-inertias method for parallel mechanisms}
\label{sec:3}
As mentioned above, there are five inertial bodies in our under-study mechanism. Such inertias generate the wrenches $\hat{\F}_{\!_{\!B0}}$, $\hat{\F}_{\!_{\!B1}}$, $\hat{\F}_{\!_{\!B3}}$, $\hat{\F}_{\!_{\!B4}}$, and $\F_{\!_{\!E}}$ expressed in their local frames.

For analyzing the interaction of these wrenches that belong to a multibody mechanism, we can use the articulated-body inertias method reported by \cite{Bib:Featherstone}. By using this method, we can assembly all inertias in a single reference frame of study to find a forward dynamics solution.

\subsection{Linear acceleration in actuator}
\label{sec:3A}
The first target is to analyze the total wrench in the actuator reference frame by means of the general expression
\begin{equation}
	\F_{\!\!_{\!B4}} \ = \ \bld{M}^{A}_{\!_{\!B4}} \dot{\bld{\nu}}_{\!_{\!B4}} + \bld{p}^{A}_{\!_{\!B4}}
	\label{eq:wrench_2e}
\end{equation}
where $\bld{M}^{A}_{\!_{\!B4}}$ and $\bld{p}^{A}_{\!_{\!B4}}$ are the cumulative spatial inertia matrix and wrench bias, respectively.

An expression for the total wrench $\F_{\!\!_{\!B4}}$, which only depends in the local wrenches $\hat{\F}_{i}$ and kinematic matrices $\bld{K}_{i}$, can be obtained following the procedure described in \ref{app:force_piston}. Now we recall the main solution (\ref{eq:99}) as
\begin{equation}
	\F_{\!\!_{\!B4}} \ = \ \bld{K}_{\!1} \widetilde{\F}_{\!\!_{\!B1}} + \bld{K}_{\!3} \hat{\F}_{\!\!_{\!B3}} + \bld{K}_{\!4} \hat{\F}_{\!\!_{\!B4}}
	\label{eq:Fb4}
\end{equation}
where we can assemble the corresponding inertias for $\hat{\F}_{\!\!_{\!B1}}$ and $\F_{\!\!_{\!E}}$ into $\widetilde{\F}_{\!\!_{\!B1}}$ by means of an inertia matrix transformation.
The wrench $\widetilde{\F}_{\!\!_{\!B1}}$ can be expressed by
\begin{equation}
	\widetilde{\F}_{\!\!_{\!B1}} \ = \ \hat{\F}_{\!\!_{\!B1}} +\mbox{Ad}_{G_{\!_{E}}^{\!_{B1}}}^{*} \F_{\!\!_{\!E}}
	\label{eq:01}
\end{equation}
where
\begin{equation}
	\F_{\!\!_{\!E}} \ = \ \hat{\F}_{\!\!_{\!E}} + \F_{\!\!_{\!ext}}
	\label{eq:54}
\end{equation}
and $\F_{\!\!_{\!ext}}$ is an external wrench, which is the force from another subsystem in a multi-DoF manipulator. For a single parallel-mechanism system, we can consider $\F_{\!\!_{\!E}}$ in its assembled form as
\begin{equation}
	\F_{\!\!_{\!E}} \ = \ \bld{M}^{A}_{\!_{\!E}} \dot{\bld{\nu}}_{\!_{\!E}} \!\!+\! \bld{p}^{A}_{\!_{\!E}}
	\label{eq:02}
\end{equation}
then by using (\ref{eq:dnu_e}), (\ref{eq:02}), and $\hat{\F}_{\!\!_{\!B1}}$ in its (\ref{eq:wrench_i}) form into (\ref{eq:01}), we can assemble $\widetilde{\F}_{\!\!_{\!B1}}$ by the expression
\begin{eqnarray}
	\widetilde{\F}_{\!\!_{\!B1}} & = & \hat{\F}_{\!\!_{\!B1}} +\mbox{Ad}_{G_{\!_{E}}^{\!_{B1}}}^{*} \F_{\!\!_{\!E}} \nonumber \\
	& = & \bld{M}_{\!_{\!B1}} \dot{\bld{\nu}}_{\!_{\!B1}} + \bld{p}_{\!_{\!B1}} + \mbox{Ad}_{G_{\!_{E}}^{\!_{B1}}}^{*} \left( \bld{M}^{A}_{\!_{\!E}} \dot{\bld{\nu}}_{\!_{\!E}} + \bld{p}^{A}_{\!_{\!E}} \right) \nonumber \\
	& = & \bld{M}_{\!_{\!B1}} \dot{\bld{\nu}}_{\!_{\!B1}} + \mbox{Ad}_{G_{\!_{E}}^{\!_{B1}}}^{*} \bld{M}^{A}_{\!_{\!E}} \mbox{Ad}_{G_{\!_{E}}^{\!_{B1}}}\dot{\bld{\nu}}_{\!_{\!B1}} + \bld{p}_{\!_{\!B1}} + \mbox{Ad}_{G_{\!_{E}}^{\!_{B1}}}^{*} \bld{p}^{A}_{\!_{\!E}} \nonumber \\
	& = & \widetilde{\bld{M}}_{\!_{\!B1}} \dot{\bld{\nu}}_{\!_{\!B1}} + \widetilde{\bld{p}}_{\!_{\!B1}} \label{eq:03}
\end{eqnarray}
where
\begin{eqnarray}
	\widetilde{\bld{M}}_{\!_{\!B1}} & = & \bld{M}_{\!_{\!B1}} + \mbox{Ad}_{G_{\!_{E}}^{\!_{B1}}}^{*} \bld{M}^{A}_{\!_{\!E}} \mbox{Ad}_{G_{\!_{E}}^{\!_{B1}}} \label{eq:41} \\
	\widetilde{\bld{p}}_{\!_{\!B1}} & = & \bld{p}_{\!_{\!B1}} + \mbox{Ad}_{G_{\!_{E}}^{\!_{B1}}}^{*} \bld{p}^{A}_{\!_{\!E}} \label{eq:42}
\end{eqnarray}

If we plug (\ref{eq:03}) and $\hat{\F}_{\!\!_{\!B3}}$ and $\hat{\F}_{\!\!_{\!B4}}$ in their (\ref{eq:wrench_i}) form into (\ref{eq:Fb4}), we can express $\F_{\!\!_{\!B4}}$ as a linear function of the spatial accelerations $\dot{\bld{\nu}}_{\!_{\!B1}}$, $\dot{\bld{\nu}}_{\!_{\!B3}}$, and $\dot{\bld{\nu}}_{\!_{\!B4}}$
\begin{eqnarray}
	\F_{\!\!_{\!B4}} & = & \bld{K}_{\!1} \widetilde{\F}_{\!\!_{\!B1}} + \bld{K}_{\!3} \hat{\F}_{\!\!_{\!B3}} + \bld{K}_{\!4} \hat{\F}_{\!\!_{\!B4}} \nonumber \\
	& = & \bld{K}_{\!1} \left( \widetilde{\bld{M}}_{\!_{\!B1}} \dot{\bld{\nu}}_{\!_{\!B1}} + \widetilde{\bld{p}}_{\!_{\!B1}} \right) + \bld{K}_{\!3} \left( \bld{M}_{\!_{\!B3}} \dot{\bld{\nu}}_{\!_{\!B3}} + \bld{p}_{\!_{\!B3}} \right) \nonumber \\
	& & + \bld{K}_{\!4} \left( \bld{M}_{\!_{\!B4}} \dot{\bld{\nu}}_{\!_{\!B4}} + \bld{p}_{\!_{\!B4}} \right) \nonumber \\
	& = & \bld{K}_{\!1} \widetilde{\bld{M}}_{\!_{\!B1}} \dot{\bld{\nu}}_{\!_{\!B1}} \!+\! \bld{K}_{\!3} \bld{M}_{\!_{\!B3}} \dot{\bld{\nu}}_{\!_{\!B3}} \!+\! \bld{K}_{\!4} \bld{M}_{\!_{\!B4}} \dot{\bld{\nu}}_{\!_{\!B4}} \!+\! \bld{\xi}^{*}_{\gamma} \label{eq:06}
\end{eqnarray}
where
\begin{equation}
	\bld{\xi}^{*}_{\gamma} \ = \ \bld{K}_{\!1} \widetilde{\bld{p}}_{\!_{\!B1}} + \bld{K}_{\!3} \bld{p}_{\!_{\!B3}} + \bld{K}_{\!4} \bld{p}_{\!_{\!B4}} \label{eq:44}
\end{equation}

For a better manipulation, we rewrite the joint accelerations defined in (\ref{eq:04}). They are now defined as the following scalar linear functions of the actuator linear acceleration $\ddot{x}$
\begin{subequations}
	\begin{eqnarray}
		\ddot{q} & = & k_4 + k_{1}\ddot{x} \label{eq:ddq_linear} \\
		\ddot{q}_{1} & = & k_5 + k_{2}\ddot{x} \label{eq:ddq1_linear} \\
		\ddot{q}_{2} & = & k_6 + k_{3}\ddot{x} \label{eq:ddq2_linear}
	\end{eqnarray}
	\label{eq:ddqall_linear}
\end{subequations}
where
\begin{equation}
	k_4 \ = \ \dot{k}_1 \dot{x} \qquad k_5 \ = \ \dot{k}_2 \dot{x} \qquad k_6 \ = \ \dot{k}_3 \dot{x}
	\label{eq:35}
\end{equation}

Moreover, we can find expressions for the spatial accelerations $\dot{\bld{\nu}}_{\!_{\!B1}}$, $\dot{\bld{\nu}}_{\!_{\!B3}}$, and $\dot{\bld{\nu}}_{\!_{\!B4}}$ as linear functions of $\ddot{x}$ and the spatial acceleration in the base reference frame $\dot{\bld{\nu}}_{\!_{\!Bc}}$. This is achieved by substituting the second-order kinematic constraints (\ref{eq:ddqall_linear}) into (\ref{eq:dnu_b1}), (\ref{eq:dnu_b3}), and (\ref{eq:dnu_b4}) as follows
\begin{subequations}
	\begin{eqnarray}
		\dot{\bld{\nu}}_{\!_{\!B1}} & = &  \mbox{Ad}_{G_{\!_{B1}}^{\!_{B0}}}\dot{\bld{\nu}}_{\!_{\!B0}} + \bld{s}_{z} k_4 + \bld{s}_{z} k_{1}\ddot{x} + \bld{c}_{\!_{B1}} \nonumber \\
		& = & \bld{\xi}_1\ddot{x} + \bld{\xi}_{3} \\
		\dot{\bld{\nu}}_{\!_{\!B3}} & = & \mbox{Ad}_{G_{\!_{B3}}^{\!_{B2}}}\dot{\bld{\nu}}_{\!_{\!B2}} + \bld{s}_{z}k_5 + \bld{s}_{z}k_{2}\ddot{x} + \bld{c}_{\!_{B3}} \nonumber \\
		& = & \bld{\xi}_4\ddot{x} + \bld{\xi}_{6} \\
		\dot{\bld{\nu}}_{\!_{\!B4}}	& = & \mbox{Ad}_{G_{\!_{B4}}^{\!_{B3}}}\dot{\bld{\nu}}_{\!_{\!B3}} + \bld{s}_{x}\ddot{x} + \bld{c}_{\!_{B4}} \nonumber \\
		& = & \mbox{Ad}_{G_{\!_{B4}}^{\!_{B3}}}\bld{\xi}_4\ddot{x} + \mbox{Ad}_{G_{\!_{B4}}^{\!_{B3}}}\bld{\xi}_{6} + \bld{s}_{x}\ddot{x} + \bld{c}_{\!_{B4}} \nonumber \\
		& = & \bld{\xi}_7\ddot{x} + \bld{\xi}_{9}
	\end{eqnarray}
	\label{eq:08}
\end{subequations}
where
\begin{subequations}
	\begin{eqnarray}
		\bld{\xi}_1 & = & \bld{s}_{z} k_{1} \\
		\bld{\xi}_2 & = & \bld{s}_{z} k_4 + \bld{c}_{\!_{B1}} \\
		\bld{\xi}_3 & = & \mbox{Ad}_{G_{\!_{B1}}^{\!_{B0}}}\dot{\bld{\nu}}_{\!_{\!B0}} + \bld{\xi}_2 \\
		\bld{\xi}_4 & = & \bld{s}_{z} k_{2} \\
		\bld{\xi}_5 & = & \bld{s}_{z}k_5 + \bld{c}_{\!_{B3}} \\
		\bld{\xi}_6 & = & \mbox{Ad}_{G_{\!_{B3}}^{\!_{B2}}}\dot{\bld{\nu}}_{\!_{\!B2}} + \bld{\xi}_5 \\
		\bld{\xi}_7 & = & \mbox{Ad}_{G_{\!_{B4}}^{\!_{B3}}}\bld{\xi}_4 + \bld{s}_{x} \\
		\bld{\xi}_8 & = & \mbox{Ad}_{G_{\!_{B4}}^{\!_{B3}}}\bld{\xi}_{5} + \bld{c}_{\!_{B4}} \\
		\bld{\xi}_9 & = & \mbox{Ad}_{G_{\!_{B4}}^{\!_{B2}}}\dot{\bld{\nu}}_{\!_{\!B2}} + \bld{\xi}_8
	\end{eqnarray}
	\label{eq:23}
\end{subequations}
where $\dot{\bld{\nu}}_{\!_{\!B0}}$ and $\dot{\bld{\nu}}_{\!_{\!B2}}$ are known from the relation (\ref{eq:07}).

Then by plugging the linear forms (\ref{eq:08}) into (\ref{eq:06}), a linear expression for $\F_{\!\!_{\!B4}}$ with respect to $\ddot{x}$ can be obtained
\begin{eqnarray}
	\F_{\!\!_{\!B4}} & = & \bld{K}_{\!1} \widetilde{\bld{M}}_{\!_{\!B1}} \dot{\bld{\nu}}_{\!_{\!B1}} + \bld{K}_{\!3} \bld{M}_{\!_{\!B3}} \dot{\bld{\nu}}_{\!_{\!B3}} + \bld{K}_{\!4} \bld{M}_{\!_{\!B4}} \dot{\bld{\nu}}_{\!_{\!B4}} + \bld{\xi}^{*}_{\gamma} \nonumber \\
	& = & \bld{K}_{\!1} \widetilde{\bld{M}}_{\!_{\!B1}} \left( \bld{\xi}_1\ddot{x} + \bld{\xi}_{3} \right) + \bld{K}_{\!3} \bld{M}_{\!_{\!B3}} \left( \bld{\xi}_4\ddot{x} + \bld{\xi}_{6} \right) \nonumber \\
	&& + \bld{K}_{\!4} \bld{M}_{\!_{\!B4}} \left( \bld{\xi}_7\ddot{x} + \bld{\xi}_{9} \right) + \bld{\xi}^{*}_{\gamma} \nonumber \\
	& = & \bld{\xi}^{*}_{\alpha}\ddot{x} + \bld{\xi}^{*}_{\beta} \label{eq:09}
\end{eqnarray}
where
\begin{eqnarray}
	\bld{\xi}^{*}_{\alpha} & = & \bld{K}_{\!1} \widetilde{\bld{M}}_{\!_{\!B1}} \bld{\xi}_1 + \bld{K}_{\!3} \bld{M}_{\!_{\!B3}} \bld{\xi}_4 + \bld{K}_{\!4} \bld{M}_{\!_{\!B4}} \bld{\xi}_7 \label{eq:43} \\
	\bld{\xi}^{*}_{\beta} & = & \bld{K}_{\!1} \widetilde{\bld{M}}_{\!_{\!B1}}\bld{\xi}_{3} + \bld{K}_{\!3} \bld{M}_{\!_{\!B3}} \bld{\xi}_{6} + \bld{K}_{\!4} \bld{M}_{\!_{\!B4}} \bld{\xi}_{9} + \bld{\xi}^{*}_{\gamma} \label{eq:25}
\end{eqnarray}

The expression (\ref{eq:09}) shows that $\F_{\!\!_{\!B4}}$ is a linear combination of the dual Lie algebra vectors $\bld{\xi}^{*}_{\alpha}$ and $\bld{\xi}^{*}_{\beta}$. Thus, $f_{\!_{B4}}^{x}$ can be computed by the screw projection presented in (\ref{eq:dyn_04})
\begin{equation*}
	f_{\!_{B4}}^{x} \ = \ \bld{s}_{x}^{\top} \F_{\!_{\!B4}} \ = \ \bld{s}_{x}^{\top} \left( \bld{\xi}^{*}_{\alpha}\ddot{x} + \bld{\xi}^{*}_{\beta} \right) \nonumber
\end{equation*}

If we define the following scalars
\begin{equation}
	\alpha \ = \ \bld{s}_{x}^{\top} \bld{\xi}^{*}_{\alpha} \qquad\quad \beta \ = \ \bld{s}_{x}^{\top} \bld{\xi}^{*}_{\beta} \label{eq:31}
\end{equation}
then the inverse dynamics is given by the scalar, analytic, and linear expression
\begin{equation}
	f_{\!_{B4}}^{x} \ = \ \alpha\ddot{x} + \beta
\end{equation}
where only $\beta$ is a function of $\dot{\bld{\nu}}_{\!_{\!Bc}}$.

Moreover, the forward dynamics is analytically given by
\begin{equation}
	\ddot{x} \ = \ \alpha^{-1}\left(f_{\!_{B4}}^{x} - \beta\right)
	\label{eq:12}
\end{equation}
which only requires to invert one scalar inertial term without increasing the algorithmic complexity. Additionally, equation (\ref{eq:12}) demonstrates that the linear acceleration in the actuator is directly proportional to the linear force acting on it.

\subsection{Assembled total wrench in base reference frame}
\label{sec:assembled}

To calculate the dynamics of a multi-DoF system (i.e., multiple closed-kinematic chains connected in a series manner), it is essential to compute the assembled expression of the total wrench exerted in the base reference frame $\Sigma_{Bc}$
\begin{equation}
	\F_{\!\!_{\!Bc}} \ = \ \bld{M}^{A}_{\!_{\!Bc}} \dot{\bld{\nu}}_{\!_{\!Bc}} + \bld{p}^{A}_{\!_{\!Bc}}
	\label{eq:wrench_bc}
\end{equation}

Thus, the objective of this section is to find expressions for the accumulative inertia matrix $\bld{M}^{A}_{\!_{\!Bc}}$ and the wrench bias $\bld{p}^{A}_{\!_{\!Bc}}$ as the independent functions of any of the accelerations in the closed chain whether they are spatial or joint accelerations.

By plugging (\ref{eq:dyn_01}) and (\ref{eq:dyn_02}) into the first and second terms of (\ref{eq:dyn_03}), respectively, an expression for the total wrench $\F_{\!\!_{\!Bc}}$, which depends only in local wrenches, is found
\begin{equation}
	\F_{\!\!_{\!Bc}} \ = \ \hat{\F}_{\!\!_{\!B0}} + \mbox{Ad}_{G_{\!_{B1}}^{\!_{Bc}}}^{*} \widetilde{\F}_{\!\!_{\!B1}} + \mbox{Ad}_{G_{\!_{B3}}^{\!_{Bc}}}^{*} \hat{\F}_{\!\!_{\!B3}} + \mbox{Ad}_{G_{\!_{B4}}^{\!_{Bc}}}^{*} \hat{\F}_{\!\!_{\!B4}}
	\label{eq:16}
\end{equation}
Note that (\ref{eq:16}) is not longer a function of the internal wrench $\F_{\!\!_{\!P}}$, such fact facilitates the following analysis.

All wrenches in (\ref{eq:16}) are expressed in their local reference frames; then the operator $\mbox{Ad}_{G_{\!_{i}}^{\!_{Bc}}}^{*}$ transforms each of these $se^{*}(3)$ elements to the reference frame $\Sigma_{Bc}$. This transformation can be denoted with a super index as
\begin{equation}
	\F^{_{(Bc)}}_{i} \ = \ \mbox{Ad}_{G_{\!_{i}}^{\!_{Bc}}}^{*} \F_{i}
\end{equation}
and (\ref{eq:16}) becomes
\begin{equation}
	\F_{\!\!_{\!Bc}} \ = \ \hat{\F}_{_{B0}} + \widetilde{\F}^{_{(Bc)}}_{_{B1}} + \hat{\F}^{_{(Bc)}}_{_{B3}} + \hat{\F}^{_{(Bc)}}_{_{B4}}
	\label{eq:22}
\end{equation}
which it is a function of the spatial accelerations $\dot{\bld{\nu}}_{\!_{\!Bc}}$, $\dot{\bld{\nu}}_{\!_{\!B1}}$, $\dot{\bld{\nu}}_{\!_{\!B3}}$, and $\dot{\bld{\nu}}_{\!_{\!B4}}$.

In order to rewrite $\F_{\!\!_{\!Bc}}$ into the assembled form (\ref{eq:wrench_bc}), it is needed to express the spatial accelerations $\dot{\bld{\nu}}_{\!_{\!B1}}$, $\dot{\bld{\nu}}_{\!_{\!B3}}$, and $\dot{\bld{\nu}}_{\!_{\!B4}}$ as linear equations of $\dot{\bld{\nu}}_{\!_{\!Bc}}$. For this, a linear form of $\ddot{x}$ with respect to $\dot{\bld{\nu}}_{\!_{\!Bc}}$ should be found.

By backward substituting the previous-section equations for the elements $\bld{\xi}_3$, $\bld{\xi}_6$, and $\bld{\xi}_9$, from (\ref{eq:23}) into (\ref{eq:25}), a linear form for $\bld{\xi}^{*}_{\beta}$ is calculated
\begin{eqnarray}
	\bld{\xi}^{*}_{\beta} & = & \bld{K}_{\!1} \widetilde{\bld{M}}_{\!_{\!B1}}\bld{\xi}_{3} + \bld{K}_{\!3} \bld{M}_{\!_{\!B3}} \bld{\xi}_{6} + \bld{K}_{\!4} \bld{M}_{\!_{\!B4}} \bld{\xi}_{9} + \bld{\xi}^{*}_{\gamma} \nonumber \\
	& = & \bld{K}_{\!1} \widetilde{\bld{M}}_{\!_{\!B1}}\!\!\left( \!\!\mbox{Ad}_{G_{\!_{B1}}^{\!_{Bc}}}\dot{\bld{\nu}}_{\!_{\!Bc}} \!\!+\! \bld{\xi}_2 \right)\! \!+\! \bld{K}_{\!3} \bld{M}_{\!_{\!B3}}\!\left( \!\mbox{Ad}_{G_{\!_{B3}}^{\!_{Bc}}}\dot{\bld{\nu}}_{\!_{\!Bc}} \!+\! \bld{\xi}_5 \!\right) \nonumber \\
	& & + \bld{K}_{\!4} \bld{M}_{\!_{\!B4}}\!\left( \mbox{Ad}_{G_{\!_{B4}}^{\!_{Bc}}}\dot{\bld{\nu}}_{\!_{\!Bc}} + \bld{\xi}_8 \right) + \bld{\xi}^{*}_{\gamma} \nonumber \\
	& = & \bld{\Psi}\dot{\bld{\nu}}_{\!_{\!Bc}} + \bld{\xi}^{*}_{\delta} \label{eq:30}
\end{eqnarray}
where
\begin{eqnarray}
	\bld{\Psi} \!\!& \!=\! &\!\! \bld{K}_{\!1} \widetilde{\bld{M}}_{\!_{\!B1}}\! \mbox{Ad}_{G_{\!_{B1}}^{\!_{Bc}}} \!\!+\! \bld{K}_{\!3} \bld{M}_{\!_{\!B3}} \mbox{Ad}_{G_{\!_{B3}}^{\!_{Bc}}} \!\!+\! \bld{K}_{\!4} \bld{M}_{\!_{\!B4}} \mbox{Ad}_{G_{\!_{B4}}^{\!_{Bc}}} \label{eq:45} \\
	\bld{\xi}^{*}_{\delta} \!\!& \!=\! &\!\! \bld{K}_{\!1} \widetilde{\bld{M}}_{\!_{\!B1}}\bld{\xi}_{2} + \bld{K}_{\!3} \bld{M}_{\!_{\!B3}} \bld{\xi}_{5} + \bld{K}_{\!4} \bld{M}_{\!_{\!B4}} \bld{\xi}_{8} + \bld{\xi}^{*}_{\gamma} \label{eq:46}
\end{eqnarray}

Moreover, by plugging (\ref{eq:30}) into (\ref{eq:31}), the scalar $\beta$ is expressed by a linear form
\begin{equation}
	\beta \ = \ \bld{s}_{x}^{\top} \bld{\xi}^{*}_{\beta} \ = \ \bld{s}_{x}^{\top} \bld{\Psi}\dot{\bld{\nu}}_{\!_{\!Bc}} + \bld{s}_{x}^{\top}\bld{\xi}^{*}_{\delta} \ = \ \bld{\psi}\dot{\bld{\nu}}_{\!_{\!Bc}} + \delta
\end{equation}
where
\begin{equation}
	\bld{\psi} \ = \ \bld{s}_{x}^{\top} \bld{\Psi} \ \in \ \real^{1\times6} \qquad \text{and} \qquad \delta \ = \ \bld{s}_{x}^{\top}\bld{\xi}^{*}_{\delta} \label{eq:47}
\end{equation}

Therefore, an additional formula for the forward dynamics (\ref{eq:12}), as a linear form of not only the actuator force $f_{\!_{B4}}^{x}$ but also the base spatial acceleration $\dot{\bld{\nu}}_{\!_{\!Bc}}$, is found 
\begin{equation}
	\ddot{x} \ = \ \alpha^{-1} \left(f_{\!_{B4}}^{x} - \bld{\psi}\dot{\bld{\nu}}_{\!_{\!Bc}} - \delta\right)
	\label{eq:33}
\end{equation}

Additionally, the spatial accelerations $\dot{\bld{\nu}}_{\!_{\!B1}}$, $\dot{\bld{\nu}}_{\!_{\!B3}}$, and $\dot{\bld{\nu}}_{\!_{\!B4}}$, from (\ref{eq:08}), can now be written as linear forms of $\dot{\bld{\nu}}_{\!_{\!Bc}}$. For this, we can use the last expression (\ref{eq:33}) for $\ddot{x}$ and (\ref{eq:23}) for $\bld{\xi}_3$, $\bld{\xi}_6$, and $\bld{\xi}_9$, respectively, as follows
\begin{subequations}
	\begin{eqnarray}
		\dot{\bld{\nu}}_{\!_{\!B1}} & = & \bld{\xi}_1 \alpha^{-1} \left(f_{\!_{B4}}^{x} \!-\! \bld{\psi}\dot{\bld{\nu}}_{\!_{\!Bc}} \!-\! \delta\right) + \mbox{Ad}_{G_{\!_{B1}}^{\!_{Bc}}}\dot{\bld{\nu}}_{\!_{\!Bc}} + \bld{\xi}_2 \nonumber \\
		& = & \bld{\Psi}_{\!_{\!B1}} \dot{\bld{\nu}}_{\!_{\!Bc}} + \bld{\xi}_{_{\!B1}} \\
		\dot{\bld{\nu}}_{\!_{\!B3}} & = & \bld{\xi}_4 \alpha^{-1} \left(f_{\!_{B4}}^{x} \!-\! \bld{\psi}\dot{\bld{\nu}}_{\!_{\!Bc}} \!-\! \delta\right) + \mbox{Ad}_{G_{\!_{B3}}^{\!_{Bc}}}\dot{\bld{\nu}}_{\!_{\!Bc}} + \bld{\xi}_5 \nonumber \\
		& = & \bld{\Psi}_{\!_{\!B3}} \dot{\bld{\nu}}_{\!_{\!Bc}} + \bld{\xi}_{_{\!B3}} \\
		\dot{\bld{\nu}}_{\!_{\!B4}}	& = & \bld{\xi}_7 \alpha^{-1} \left(f_{\!_{B4}}^{x} \!-\! \bld{\psi}\dot{\bld{\nu}}_{\!_{\!Bc}} \!-\! \delta\right) + \mbox{Ad}_{G_{\!_{B4}}^{\!_{Bc}}}\dot{\bld{\nu}}_{\!_{\!Bc}} + \bld{\xi}_8 \nonumber \\
		& = & \bld{\Psi}_{\!_{\!B4}} \dot{\bld{\nu}}_{\!_{\!Bc}} + \bld{\xi}_{_{\!B4}}
	\end{eqnarray}
	\label{eq:34}
\end{subequations}
where
\begin{subequations}
	\begin{eqnarray}
		\bld{\Psi}_{\!_{\!B1}} & = & \mbox{Ad}_{G_{\!_{B1}}^{\!_{Bc}}} - \bld{\xi}_1 \alpha^{-1} \bld{\psi} \\
		\bld{\Psi}_{\!_{\!B3}} & = & \mbox{Ad}_{G_{\!_{B3}}^{\!_{Bc}}} - \bld{\xi}_4 \alpha^{-1} \bld{\psi} \\
		\bld{\Psi}_{\!_{\!B4}} & = & \mbox{Ad}_{G_{\!_{B4}}^{\!_{Bc}}} - \bld{\xi}_7 \alpha^{-1} \bld{\psi} \\
		\bld{\xi}_{_{\!B1}} & = & \bld{\xi}_1 \alpha^{-1} \left(f_{\!_{B4}}^{x} - \delta\right) + \bld{\xi}_2 \\
		\bld{\xi}_{_{\!B3}} & = & \bld{\xi}_4 \alpha^{-1} \left(f_{\!_{B4}}^{x} - \delta\right) + \bld{\xi}_5 \\
		\bld{\xi}_{_{\!B4}} & = & \bld{\xi}_7 \alpha^{-1} \left(f_{\!_{B4}}^{x} - \delta\right) + \bld{\xi}_8
	\end{eqnarray}
	\label{eq:48}
\end{subequations}

Since the wrench $\F_{\!\!_{\!Bc}}$ in equation (\ref{eq:22}) is the sum up of wrenches, we can insert the wrench form (\ref{eq:03}) for $\widetilde{\F}^{_{(Bc)}}_{\!\!_{\!B1}}$ and the general form (\ref{eq:wrench_i}) for $\hat{\F}_{\!\!_{\!B0}}$, $\hat{\F}^{_{(Bc)}}_{\!\!_{\!B3}}$, and $\hat{\F}^{_{(Bc)}}_{\!\!_{\!B4}}$. This procedure remains in a function for $\F_{\!\!_{\!Bc}}$ that depends on the spatial accelerations $\dot{\bld{\nu}}_{\!_{\!Bc}}$, $\dot{\bld{\nu}}_{\!_{\!B1}}$, $\dot{\bld{\nu}}_{\!_{\!B3}}$, and $\dot{\bld{\nu}}_{\!_{\!B4}}$. Then by replacing $\dot{\bld{\nu}}_{\!_{\!B1}}$, $\dot{\bld{\nu}}_{\!_{\!B3}}$, and $\dot{\bld{\nu}}_{\!_{\!B4}}$, from (\ref{eq:34}), it results in a function that only depends on $\dot{\bld{\nu}}_{\!_{\!Bc}}$. We sketch this procedure for each wrench in (\ref{eq:22}) as follows
\begin{eqnarray}
	\hat{\F}_{\!\!_{\!B0}} & = & \bld{M}_{\!_{\!B0}} \dot{\bld{\nu}}_{\!_{\!Bc}} + \bld{p}_{\!_{\!B0}} \nonumber \\
	\widetilde{\F}^{_{(Bc)}}_{\!\!_{\!B1}} & = & \mbox{Ad}_{G_{\!_{B1}}^{\!_{Bc}}}^{*} \left( \widetilde{\bld{M}}_{\!_{\!B1}} \left( \bld{\Psi}_{\!_{\!B1}} \dot{\bld{\nu}}_{\!_{\!Bc}} + \bld{\xi}_{_{\!B1}} \right) + \widetilde{\bld{p}}_{\!_{\!B1}} \right) \nonumber \\
	\hat{\F}^{_{(Bc)}}_{\!\!_{\!B3}} & = & \mbox{Ad}_{G_{\!_{B3}}^{\!_{Bc}}}^{*} \left( \bld{M}_{\!_{\!B3}} \left( \bld{\Psi}_{\!_{\!B3}} \dot{\bld{\nu}}_{\!_{\!Bc}} + \bld{\xi}_{_{\!B3}} \right) + \bld{p}_{\!_{\!B3}} \right) \nonumber \\
	\hat{\F}^{_{(Bc)}}_{\!\!_{\!B4}} & = & \mbox{Ad}_{G_{\!_{B4}}^{\!_{Bc}}}^{*} \left( \bld{M}_{\!_{\!B4}} \left( \bld{\Psi}_{\!_{\!B4}} \dot{\bld{\nu}}_{\!_{\!Bc}} + \bld{\xi}_{_{\!B4}} \right) + \bld{p}_{\!_{\!B4}} \right) \nonumber
\end{eqnarray}
and by adding them, the cumulative inertial matrix $\bld{M}^{A}_{\!_{\!Bc}}$ and cumulative wrench bias $\bld{p}^{A}_{\!_{\!Bc}}$ in equation (\ref{eq:wrench_bc}) can be computed by means of the expressions
\begin{subequations}
	\begin{eqnarray}
		\bld{M}^{A}_{\!_{\!Bc}} & = & \bld{M}_{\!_{\!B0}} + \mbox{Ad}_{G_{\!_{B1}}^{\!_{Bc}}}^{*} \widetilde{\bld{M}}_{\!_{\!B1}} \bld{\Psi}_{\!_{\!B1}} + \mbox{Ad}_{G_{\!_{B3}}^{\!_{Bc}}}^{*} \bld{M}_{\!_{\!B3}} \bld{\Psi}_{\!_{\!B3}} \nonumber \\
		&  &+ \mbox{Ad}_{G_{\!_{B4}}^{\!_{Bc}}}^{*} \bld{M}_{\!_{\!B4}} \bld{\Psi}_{\!_{\!B4}} \\
		\bld{p}^{A}_{\!_{\!Bc}} & = & \bld{p}_{\!_{\!B0}} + \mbox{Ad}_{G_{\!_{B1}}^{\!_{Bc}}}^{*} \left( \widetilde{\bld{M}}_{\!_{\!B1}} \bld{\xi}_{_{\!B1}} + \widetilde{\bld{p}}_{\!_{\!B1}} \right) \\
		&  & + \mbox{Ad}_{G_{\!_{B3}}^{\!_{Bc}}}^{*} \!\!\left(\! \bld{M}_{\!_{\!B3}} \bld{\xi}_{_{\!B3}} \!\!+\! \bld{p}_{\!_{\!B3}} \right) + \mbox{Ad}_{G_{\!_{B4}}^{\!_{Bc}}}^{*} \!\!\left( \bld{M}_{\!_{\!B4}} \bld{\xi}_{_{\!B4}} \!\!+\! \bld{p}_{\!_{\!B4}} \right) \nonumber
	\end{eqnarray}
	\label{eq:49}
\end{subequations}
\section{Forward-dynamics recursive algorithm for parallel-serial manipulators}
\label{sec:manipulator}
A parallel-serial manipulator is defined as a multi-DoF serial kinematic chain of 1-DoF mechanical modules. Such modules can be a parallel mechanism, as described in Section \ref{sec:dynamics_parallel}, or a single-DoF-joint serial mechanism, whether prismatic or revolute. In this section, we define the dynamics of serial mechanism and then generate a complete algorithm based on the assembly method, which can solve the forward dynamics of the manipulator.

Let us recall here the main equations of the ABA \cite{Bib:Featherstone} for single-DoF serial mechanism. This kinematic chain has three reference frames: $\Sigma_{Bc}$, $\Sigma_{B4}$ and $\Sigma_{E}$. Assuming the assembly components of $\F_{\!\!_{\!E}}$ are known (i.e., $\bld{M}^{A}_{\!_{\!E}}$ and $\bld{p}^{A}_{\!_{\!E}}$ in (\ref{eq:02})), they can be backwardly transformed to $\Sigma_{B4}$ by
\begin{subequations}
	\begin{eqnarray}
		\bld{M}^{A}_{\!_{\!B4}} & = & \bld{M}_{\!_{\!B4}} + \mbox{Ad}_{G_{\!_{E}}^{\!_{B4}}}^{*} \bld{M}^{A}_{\!_{\!E}} \mbox{Ad}_{G_{\!_{E}}^{\!_{B4}}} \\
		\bld{p}^{A}_{\!_{\!B4}} & = & \bld{p}_{\!_{\!B4}} + \mbox{Ad}_{G_{\!_{E}}^{\!_{B4}}}^{*} \bld{p}^{A}_{\!_{\!E}}
	\end{eqnarray}
	\label{eq:52}
\end{subequations}
Then, by defining the next auxiliary variables
\begin{equation}
	\bld{\psi} \ = \ \bld{s}_{\!_{B4}}^{\top} \bld{M}^{A}_{\!_{\!B4}} \qquad \alpha \ = \ \bld{\psi} \bld{s}_{\!_{B4}} \qquad \bld{u} \ = \ f_{\!_{B4}} - \bld{s}_{\!_{B4}}^{\top} \bld{p}^{A}_{\!_{\!B4}} \label{eq:50}
\end{equation}
the assembly components, $\bld{M}^{A}_{\!_{\!Bc}}$ and $\bld{p}^{A}_{\!_{\!Bc}}$, of $\F_{\!\!_{\!Bc}}$ in (\ref{eq:wrench_bc}) are computed as
\begin{subequations}
	\begin{eqnarray}
		\bld{M}^{a}_{\!_{\!B4}} & = & \bld{M}^{A}_{\!_{\!B4}} - \bld{\psi}^{\top} \alpha^{-1}\bld{\psi} \label{eq:24} \\
		\bld{p}^{a}_{\!_{\!B4}} & = &\bld{p}^{A}_{\!_{\!B4}} + \bld{M}^{a}_{\!_{\!B4}}\bld{c}_{\!_{B4}} + \bld{u}\alpha^{-1}\bld{\psi}^{\top} \label{eq:25a} \\
		\bld{M}^{A}_{\!_{\!Bc}} & = & \bld{M}_{\!_{\!Bc}} + \mbox{Ad}_{G_{\!_{B4}}^{\!_{Bc}}}^{*} \bld{M}^{a}_{\!_{\!B4}} \mbox{Ad}_{G_{\!_{B4}}^{\!_{Bc}}} \\
		\bld{p}^{A}_{\!_{\!Bc}} & = & \bld{p}_{\!_{\!Bc}} + \mbox{Ad}_{G_{\!_{B4}}^{\!_{Bc}}}^{*} \bld{p}^{a}_{\!_{\!B4}}
	\end{eqnarray}
	\label{eq:51}
\end{subequations}
where $f_{\!_{B4}}$ is the force/torque in the actuator, $\bld{\psi} \in \real^{1\times6}$, and $\bld{s}_{\!_{B4}}$ is the single screw vector of the serial mechanism.

\begin{algorithm}[h!]
	\small{
		\DontPrintSemicolon 
		\KwIn{$ \ \x \quad \dot{\x} \quad \bld{f}$}
		\tcc{Initialization}
		$\bld{\nu}_0 \leftarrow \bld{0} \qquad \dot{\bld{\nu}}_0 \leftarrow-\dot{\bld{\nu}}_{g}$ \;
		\tcc{First forward recursion}
		\For{$\kappa=1$ \KwTo $n$}{
			\If{parallel mechanism}{
				Solve $k_{j\kappa} \ \forall j=1,\cdots,6$ from (\ref{eq:k1_3}) and (\ref{eq:35}). \;
				Solve bias $\bld{c}_{\!_{B1\kappa}}$, $\bld{c}_{\!_{B3\kappa}}$, and $\bld{c}_{\!_{B4\kappa}}$ with (\ref{eq:37}). \;
				Solve bias $\bld{p}_{\!_{\!B1\kappa}}$, $\bld{p}_{\!_{\!B3\kappa}}$, $\bld{p}_{\!_{\!B4\kappa}}$, and $\bld{p}_{\!_{\!E\kappa}}$ with (\ref{eq:38}). \;
				Solve elements $\bld{\xi}_{j\kappa} \ \forall j=1,2,4,5,7,8$ with (\ref{eq:23}). \;
			}
			\If{serial mechanism}{
				Solve bias $\bld{c}_{\!_{B4\kappa}}$ with (\ref{eq:37}). \;
				Solve bias $\bld{p}_{\!_{\!Bc\kappa}}$, $\bld{p}_{\!_{\!B4\kappa}}$, and $\bld{p}_{\!_{\!E\kappa}}$ with (\ref{eq:38}).
			}			
		}
		\tcc{Backward recursion}
		\For{$\kappa=n$ \KwTo $1$}{
			\If{parallel mechanism}{
				Retrieve $\bld{K}_{\!1\kappa}$, $\bld{K}_{\!3\kappa}$, and $\bld{K}_{\!4\kappa}$ from (\ref{eq:39} - \ref{eq:40}). \;			
				Solve $\bld{\xi}^{*}_{\alpha\kappa}$ and $\alpha_{\kappa}$ with (\ref{eq:41}), (\ref{eq:43}), and (\ref{eq:31}). \;
				Solve $\bld{\xi}^{*}_{\gamma\kappa}$ with (\ref{eq:42}) and (\ref{eq:44}). \;
				Compute $\bld{\Psi}_{\kappa}$ and $\bld{\xi}^{*}_{\delta\kappa}$ with (\ref{eq:45}) and (\ref{eq:46}). \;
				Compute $\bld{\psi}_{\kappa}$ and $\delta_{\kappa}$ with (\ref{eq:47}). \;
				Solve $\!\bld{\Psi}_{\!_{\!B1\kappa}}$, $\!\!\bld{\Psi}_{\!_{\!B3\kappa}}$, $\!\!\bld{\Psi}_{\!_{\!B4\kappa}}$, $\!\bld{\xi}_{_{\!B1\kappa}}$, $\!\bld{\xi}_{_{\!B3\kappa}}$, and $\!\bld{\xi}_{_{\!B4\kappa}}$ with (\ref{eq:48}). \;
				Solve $\bld{M}^{A}_{\!_{\!Bc\,\kappa}}$ and $\bld{p}^{A}_{\!_{\!Bc\,\kappa}}$ with (\ref{eq:49}).
			}
			\If{serial mechanism}{
				Solve $\bld{M}^{A}_{\!_{\!B4\kappa}}$ and $\bld{p}^{A}_{\!_{\!B4\kappa}}$ with (\ref{eq:52}). \;
				Solve $\bld{\psi}_{\kappa}$, $\alpha_{\kappa}$, and $\bld{u}_{\kappa}$ with (\ref{eq:50}). \;
				Solve $\bld{M}^{a}_{\!_{\!B4\kappa}}$, $\bld{p}^{a}_{\!_{\!B4\kappa}}$, $\bld{M}^{A}_{\!_{\!Bc\,\kappa}}$, and $\bld{p}^{A}_{\!_{\!Bc\,\kappa}}$ with (\ref{eq:51}). \;
			}
			Use $\bld{M}^{A}_{\!_{\!Bc\,\kappa}}$ and $\bld{p}^{A}_{\!_{\!Bc\,\kappa}}$ as components of $\F_{\!\!_{\!ext, \kappa-1}}$ in (\ref{eq:54}). \;
		}
		\tcc{Second forward recursion}
		\For{$i=\kappa$ \KwTo $n$}{
			\If{parallel mechanism}{
				Compute $\dot{\bld{\nu}}_{\!_{\!Bc\kappa}}$ from previous spatial accelerations. \;
				Solve dynamics $\ddot{x}_{\kappa}$ with linear form (\ref{eq:33}). \;				
				Solve all spatial accelerations with (\ref{eq:04}) and (\ref{eq:53}). \;				
			}		
			\If{serial mechanism}{
				Compute $\dot{\bld{\nu}}_{\!_{\!B4\kappa}}^{a} = \mbox{Ad}_{G_{\!_{B4}}^{\!_{Bc}}}\dot{\bld{\nu}}_{\!_{\!Bc\kappa}} + \bld{c}_{\!_{B4\kappa}}$ \;
				Solve dynamics $\ddot{x}_{\kappa} = \alpha_{\kappa}^{-1}\left( \bld{u}_{\kappa}-\bld{\psi}_{\kappa}\dot{\bld{\nu}}_{\!_{\!B4\kappa}}^{a} \right)$ \;				
				Update $\dot{\bld{\nu}}_{\!_{\!B4\kappa}} = \dot{\bld{\nu}}_{\!_{\!B4\kappa}}^{a} + \bld{s}_{\!_{B4\kappa}}\ddot{x}_{\kappa}$ \;
			}
		}
		{\Return{$\bld{\ddot{\x}} \leftarrow [\ddot{x}_1 \ \ddot{x}_2 \ \cdots \ \ddot{x}_n]^{\top}$}\;}
	}
	\caption{\small{{\textbf {ParallelSerialManipulatorDynamics}}}}
	\label{algo:ABA}
\end{algorithm}

The previous sections have used notation only for the local mechanical module, whether a parallel or serial mechanism. Now, since the manipulator is a concatenation of $n$ mechanical modules, we refer to the $\kappa$-th mechanism by adding an extra sub index to all aforementioned equations.

Therefore, the forward dynamics of a $n$-DoF manipulator (i.e., it contains $n$ mechanical modules), can be computed by \textbf{Algorithm} \textbf{\ref{algo:ABA}}, where $\bld{f}$ is the input forces vector, $\x$, $\dot{\x}$, and $\ddot{\x}$ are the position, velocity, and accelerations of the actuators, and $\dot{\bld{\nu}}_{g}$ is the acceleration due to gravity. \textbf{Algorithm} \textbf{\ref{algo:ABA}} is composed by three recursive loops with conditionals for parallel and serial mechanisms. The dynamical coupling among the modules is described in line 24 assuming that the base-frame wrench of module $\kappa$ is the input wrench of module $\kappa-1$.

\section{Results}
\label{sec:results}
\begin{figure}[h!]
	\centering
	\begin{subfigure}[b]{0.122\textwidth}
		\includegraphics[trim={0.0cm 0.0cm 0.0cm 0.0cm},clip,width=\textwidth]{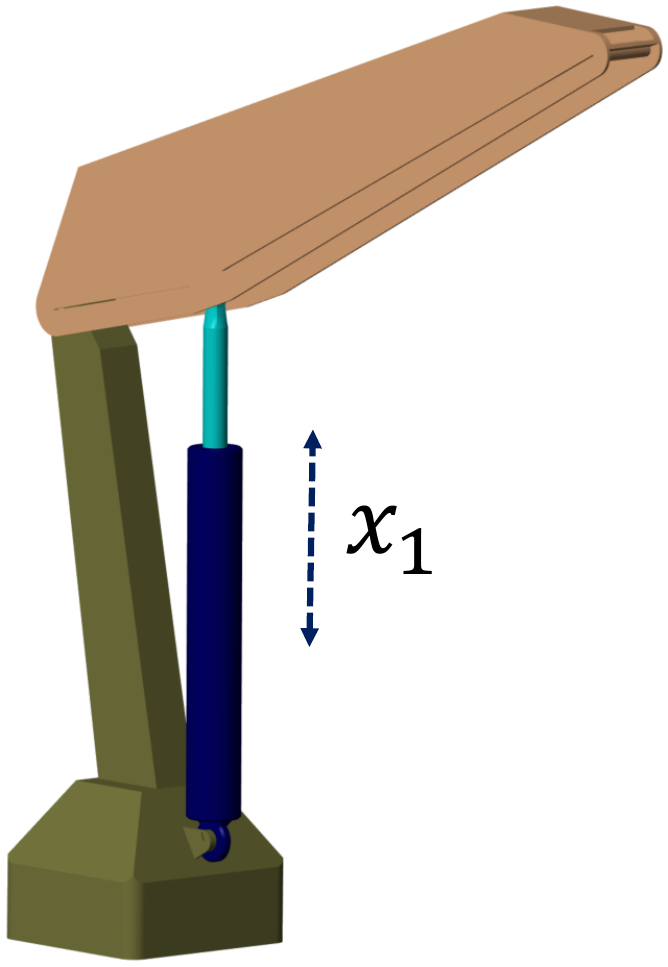}
		\vspace*{-0.5cm}
		\caption{Closed loop.}
		\label{fig:R_a}
	\end{subfigure}
	\hspace*{0.5cm}
	\begin{subfigure}[b]{0.3\textwidth}
		\includegraphics[trim={0.0cm 0.0cm 0.0cm 0.0cm},clip,width=\textwidth]{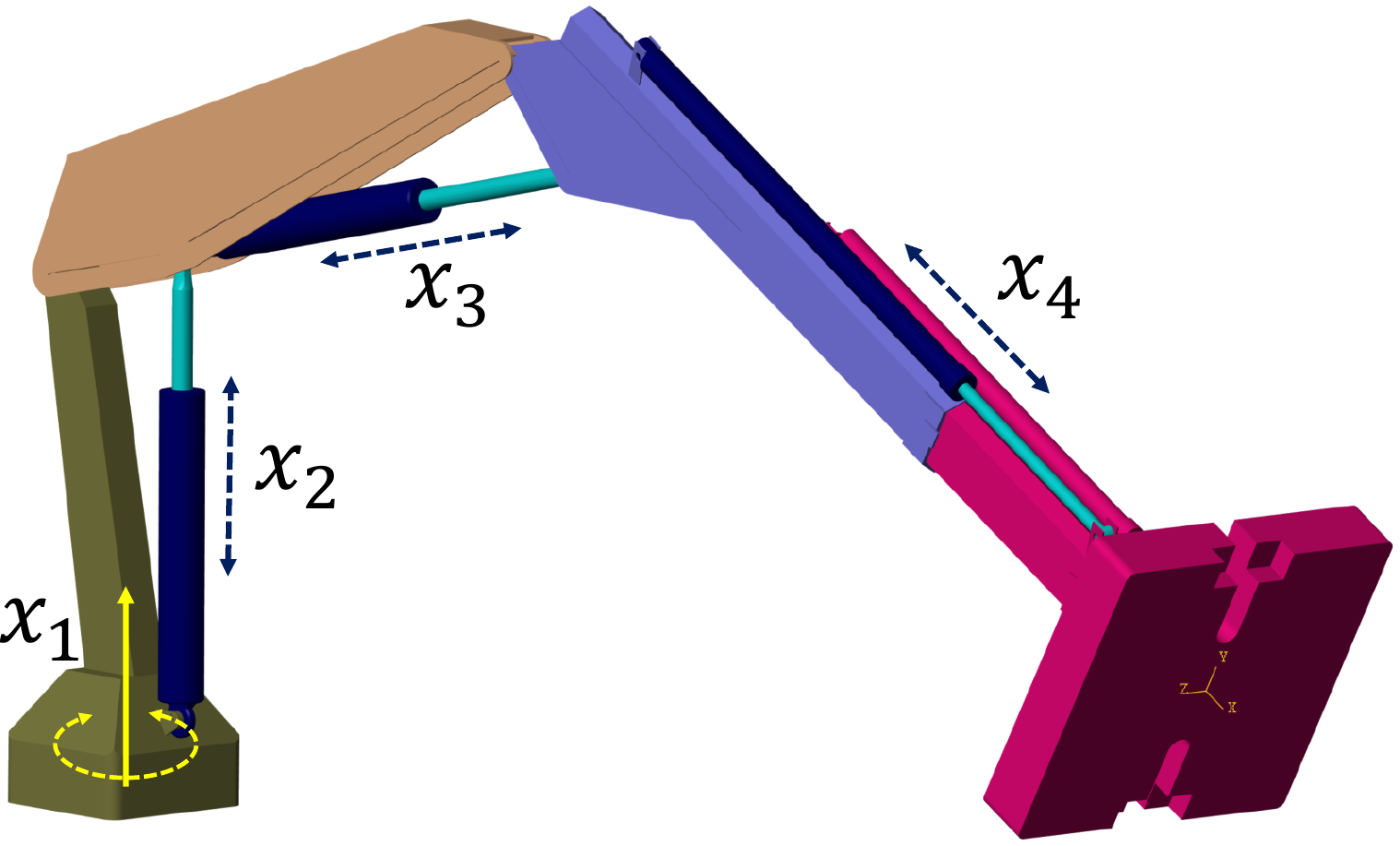}
		\vspace*{-0.5cm}
		\caption{Parallel-serial manipulator.}
		\label{fig:R_b}
	\end{subfigure}
	\caption{{\small {\bf Mechanisms under study that contain kinematics closed loops.} (a) shows a 1-DoF parallel mechanism while (b) is a 4-DoF manipulator containing two closed loops and two single-joint DoF.}}
	\label{fig:R_}
\end{figure}

\begin{figure}[h!]
	\centering
	\begin{subfigure}[t]{\columnwidth}
		\includegraphics[trim={0.05cm 21.5cm 0.2cm 0.4cm},clip,width=\columnwidth]{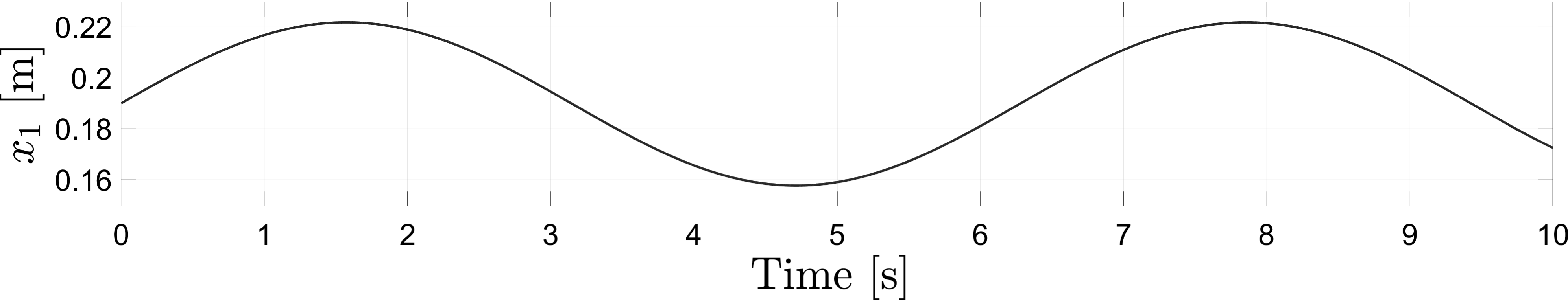}
		\vspace*{-0.6cm}
		\caption{{\footnotesize Linear actuator position $x_1$ through time.}}
	\end{subfigure}
	
	\vspace*{0.2cm}
	\begin{subfigure}[t]{\columnwidth}
		\includegraphics[trim={0.0cm 20.5cm 0.08cm 0.05cm},clip,width=\columnwidth]{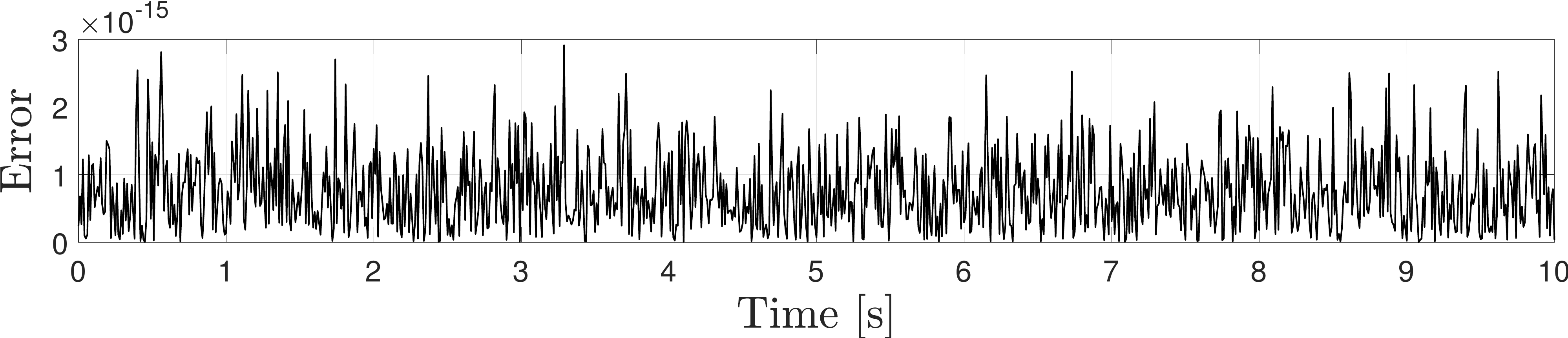}
		\vspace*{-0.6cm}
		\caption{{\footnotesize Errors of our analytic FD as the inverse function of the analytic ID, see (\ref{eq:id}).}}
	\end{subfigure}
	
	\vspace*{0.2cm}
	\begin{subfigure}[t]{\columnwidth}
		\includegraphics[trim={0.0cm 20.5cm 0.34cm 0.02cm},clip,width=\columnwidth]{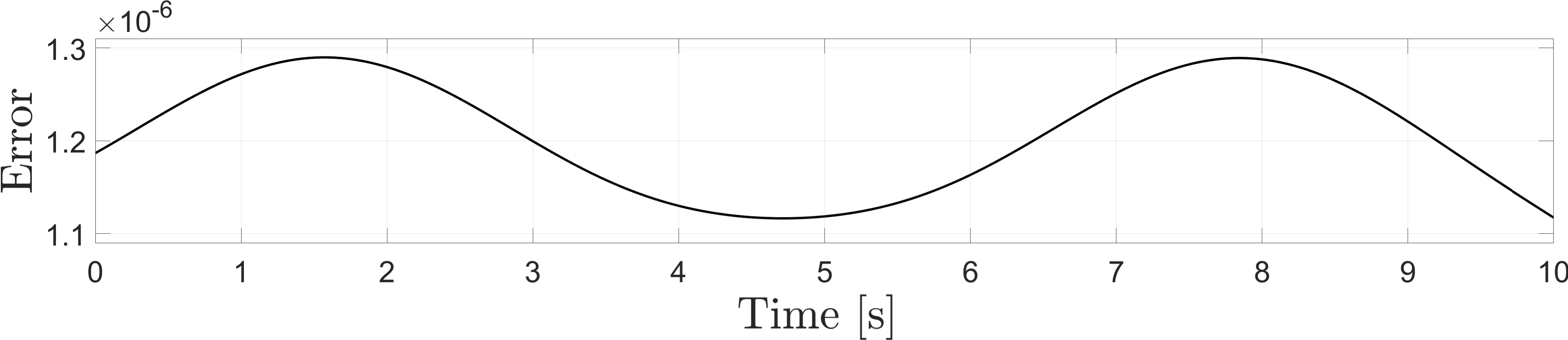}
		\vspace*{-0.6cm}
		\caption{{\footnotesize Errors among our analytic FD and Simscape physics engine.}}
	\end{subfigure}
	\vspace*{0.0cm}
	\caption{{\small Validation of our analytical forward dynamics solution for a single parallel mechanims, shown in Figure \ref{fig:R_} (a). }}
	\label{fig:R_results}
\end{figure}

We first analyze and validate the mathematical framework developed in Section \ref{sec:3A} for solving the forward dynamics of a single 1-DoF parallel mechanism of total weight 415 [$kg$]. This is depicted in Figure \ref{fig:R_} (a), and its kinematics is presented in Section \ref{sec:dynamics_parallel}. The numerical validation has been performed in M{\footnotesize ATLAB} software and Simscape toolbox for comparing our analytical solution against this physics engine simulator. The simulation was performed during 10 [$s$] of motion with sinusoidal input functions. Figure \ref{fig:R_results} (a) shows the linear actuator position $\x_1$ over time.

An analytic inverse-dynamics solution was presented in \cite{petrovic2022mathematical} as $\bld{f}=\bld{I\!D}(\bld{x},\dot{\bld{x}},\ddot{\bld{x}})$. Then, our \textbf{Algorithm} \textbf{\ref{algo:ABA}} represents the analytic forward-dynamics solution $\ddot{\bld{x}}=\bld{F\!D}(\bld{x},\dot{\bld{x}},\bld{f})$. To validate that our function $\bld{F\!D}(\cdot)$ is in fact the inverse of $\bld{I\!D}(\cdot)$, we have performed the following simulation
\begin{equation}
	\ddot{\bld{x}}_{{\scriptsize \mbox{out}}} \ = \ \bld{F\!D}\left(\bld{x}, \ \dot{\bld{x}}, \ \bld{I\!D}\left( \bld{x}, \ \dot{\bld{x}}, \ \ddot{\bld{x}}_{{\scriptsize \mbox{in}}} \right)\right)
	\label{eq:89}
\end{equation}
where $\ddot{\bld{x}}_{{\scriptsize \mbox{in}}}$ follows sinusoidal behavior, and $\ddot{\bld{x}}_{{\scriptsize \mbox{out}}}$ is the retrieved actuator acceleration. Thus, the absolute difference between $\ddot{\bld{x}}_{{\scriptsize \mbox{in}}}$ and $\ddot{\bld{x}}_{{\scriptsize \mbox{out}}}$ is depicted in Figure \ref{fig:R_results} (b), where we have obtained an average error of 7.66e-16 [$m/s^{2}$] in actuator acceleration accuracy. This error represents 1.25e-11\% of the absolute acceleration.

Additionally, we simulate the forward dynamics of this mechanism in Simscape and compare the output against our analytical solution. The error among both solutions is depicted in Figure \ref{fig:R_results} (c) with an average of 1.20e-6 [$m/s^{2}$], which represents around 0.02\% of the actuator acceleration.

However, the whole \textbf{Algorithm} \textbf{\ref{algo:ABA}} is validated with a 4-DoF heavy-duty manipulator model of total weight 1,063 [$kg$]. This system is shown in Figure \ref{fig:R_} (b) and contains a revolute joint in the base, two concatenated closed-loop kinematic chains, and finally a prismatic joint. Note that for both closed loops, we are using the procedure developed in Section \ref{sec:assembled}, since they require the backward projection of the assembled inertias.

\begin{figure}[h!]
	\centering
	\begin{subfigure}[t]{\columnwidth}
		\includegraphics[trim={0.2cm 21cm 0.2cm 0.6cm},clip,width=\columnwidth]{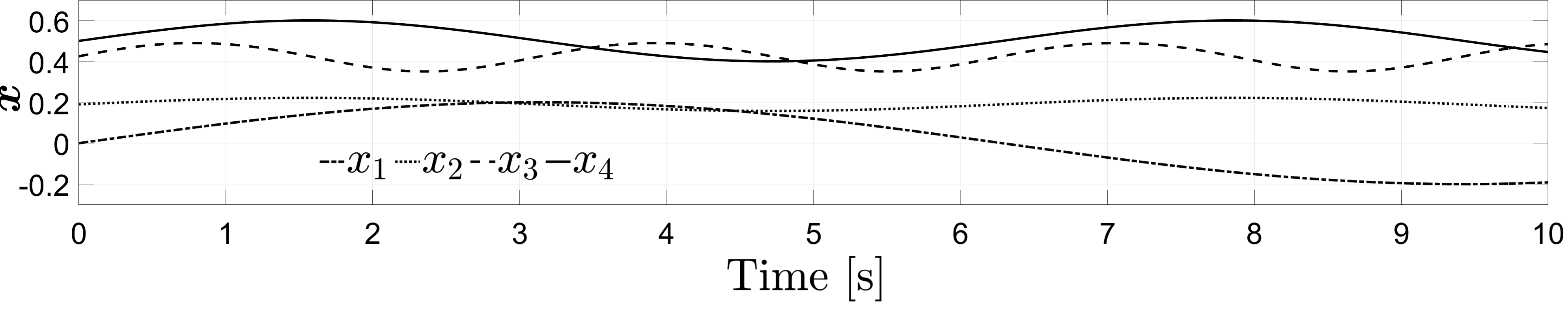}
		\vspace*{-0.6cm}
		\caption{{\footnotesize Actuators' positions ($\bld{x}$) through time.}}
	\end{subfigure}
	
	\vspace*{0.2cm}
	\begin{subfigure}[t]{\columnwidth}
		\includegraphics[trim={0.38cm 20cm 0.2cm 0.1cm},clip,width=\columnwidth]{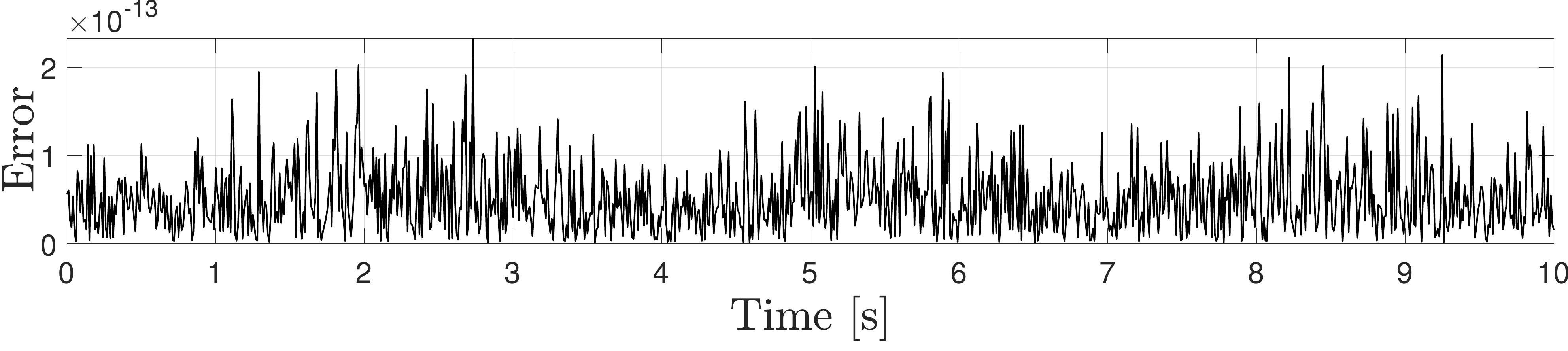}
		\vspace*{-0.6cm}
		\caption{{\footnotesize Errors of our analytic FD as the inverse function of the analytic ID, see (\ref{eq:id}).}}
	\end{subfigure}
	
	\vspace*{0.2cm}
	\begin{subfigure}[t]{\columnwidth}
		\includegraphics[trim={0.1cm 20cm 0.33cm 0.06cm},clip,width=\columnwidth]{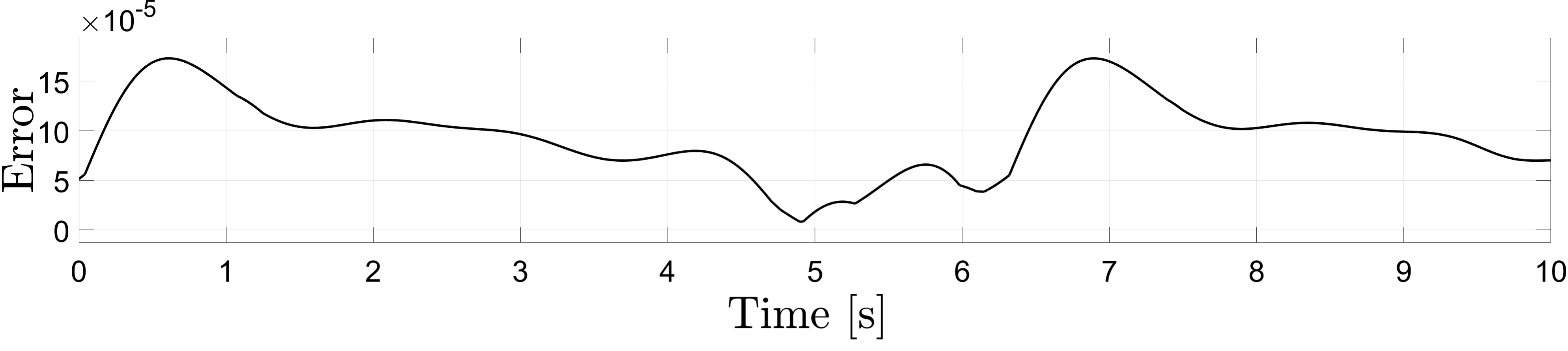}
		\vspace*{-0.6cm}
		\caption{{\footnotesize Errors among our analytic FD and Simscape physics engine.}}
	\end{subfigure}
	\vspace*{0.0cm}
	\caption{{\small Validation of our analytical forward dynamics solution for a paralle-serial manipulator, shown in Figure \ref{fig:R_} (b). }}
	\label{fig:RRRP_results}
\end{figure}

Under sinusoidal joint-position inputs, depicted in Figure \ref{fig:RRRP_results} (a), for a lapse of 10 [$s$], we test our solution as the inverse function of analytic $\bld{I\!D}(\cdot)$. By using the expression (\ref{eq:89}), we measure the error as the sum of absolute difference between $\ddot{\bld{x}}_{{\scriptsize \mbox{in}}}$ and $\ddot{\bld{x}}_{{\scriptsize \mbox{out}}}$. This error through time is shown in Figure \ref{fig:RRRP_results} (b), and it has an average of 5.21e-14 that represents 2.42e-11\% of the total acceleration in actuators' joints.

We also compare our analytic forward dynamics solution with the Simscape simulator for the parallel-serial manipulator. The accuracy is shown by the error among solutions in Figure \ref{fig:RRRP_results} (c). Such an error has an average of 9.59e-5 and represents 0.04\% of the total sum of acceleration in the actuators' joints.

\section{Conclusions}
\label{sec:conclusions}

We have generated a new recursive formulation to analytically compute the following expressions:
\begin{itemize}
	\item An equation for the total wrench in the actuator's reference frame $\F_{\!\!_{\!B4}}$, see (\ref{eq:99}), which does not depend on internal closed-loop forces.
	
	\item  The forward dynamics of a parallel mechanism, see (\ref{eq:12}), for computing the linear-actuator acceleration $\ddot{x}$ as a linear, scalar, and exact function of the force acting on it $f_{\!_{B4}}^{x}$. The efficiency of this formula is demonstrated by using the mechanism depicted in Figure \ref{fig:R_} (a) and achieving an accuracy of 1.25e-11\% average error when comparing against analytical solution (\ref{eq:89}) and 0.02\% when comparing against the Simscape physics engine.
	
	\item By applying the assembly inertias method, a solution for the forward dynamics of the parallel mechanism is presented as a linear form of not only the actuator force $f_{\!_{B4}}^{x}$ but also the base reference-frame spatial acceleration $\dot{\bld{\nu}}_{\!_{\!Bc}}$, see (\ref{eq:33}). Following the assembly methodology, we also present how to compute the cumulative inertia matrix $\bld{M}^{A}_{\!_{\!Bc}}$ and wrench bias $\bld{p}^{A}_{\!_{\!Bc}}$ acting on the base reference-frame, see (\ref{eq:49}). This allows for the computation of the forward dynamics of the multi-DoF mechanism and its efficiency is shown with the manipulator depicted in Figure \ref{fig:R_} (b). We obtain an accuracy of 2.42e-11\% when comparing against an analytic solution and 0.04\% against Simscape simulator. For the inverse dynamics problem and using the same manipulator an inputs, \cite{petrovic2022mathematical} reported an average error of 0.2\% against Simscape.
\end{itemize}

\textbf{Algorithm} \textbf{\ref{algo:ABA}} encapsulates all these expressions, and it is suitable for dealing with $n$-DoF parallel-serial manipulators while holding an $\mathcal{O}(n)$ algorithmic complexity.

\appendix

%
\section{Analytic inverse dynamics}
\label{app:inv_dyn}
All wrenches described in Section \ref{sec:dynamics_parallel} have the structure
\begin{equation}
	\F_{i} \ = \ \begin{bmatrix}
		f_{i}^{x} &
		f_{i}^{y} &
		f_{i}^{z} &
		m_{i}^{x} &
		m_{i}^{y} &
		m_{i}^{z}
	\end{bmatrix}^{\top}
\end{equation}
where $f$ and $m$ elements are the scalar linear forces and angular moments, respectively, acting on the indicated axes. 

From \cite{zhu2010virtual}, we adopt the zero-torque assumption for passive joints in the closed kinematic chain as
\begin{subequations}
	\begin{eqnarray}
		\bld{s}_{z}^{\top} \F_{\!_{\!B1}} & = & 0 \\
		\bld{s}_{z}^{\top} \F_{\!_{\!P}} & = & 0 \\
		\bld{s}_{z}^{\top} \F_{\!_{\!B3}} & = & 0 
	\end{eqnarray}
	\label{eq:assum_01}
\end{subequations}
and we also include the following assumptions from \cite{petrovic2022mathematical}
\begin{subequations}
	\begin{eqnarray}
		\F_{\!_{\!B1}} & = & \begin{bmatrix}
			f_{\!_{B1}}^{x} & f_{\!_{B1}}^{y} & 0 & 0 & 0 & 0
		\end{bmatrix}^{\top} \\
		\F_{\!_{\!P}} & = & \begin{bmatrix}
			f_{\!_{P}}^{x} & f_{\!_{P}}^{y} & 0 & 0 & 0 & 0
		\end{bmatrix}^{\top} \\
		\F_{\!_{\!B3}} & = & \begin{bmatrix}
			f_{\!_{B3}}^{x} & f_{\!_{B3}}^{y} & 0 & 0 & 0 & 0
		\end{bmatrix}^{\top} \\
		\F_{\!_{\!B4}} & = & \begin{bmatrix}
			f_{\!_{B4}}^{x} & f_{\!_{B4}}^{y} & 0 & 0 & 0 & m_{\!_{B4}}^{z}
		\end{bmatrix}^{\top} \label{eq:56}
	\end{eqnarray}
	\label{eq:assum_02}
\end{subequations}

By plugging the assumptions (\ref{eq:assum_01}) and (\ref{eq:assum_02}) into the equations from (\ref{eq:dyn_01}) to (\ref{eq:dyn_04}), as \cite{petrovic2022mathematical} suggests, we can arrive to the next $7\times7$ linear system
\begin{equation}
	\bld{A}\F_{\!\psi} \ = \ \hat{\F}_{\!\psi}
	\label{eq:linSys}
\end{equation}
where
\begin{eqnarray}
	\F_{\!\psi} & \hspace*{-0.1cm}=\hspace*{-0.1cm} & \begin{bmatrix}
		f_{\!_{P}}^{x} &
		f_{\!_{P}}^{y} &
		f_{\!_{B4}}^{x} &
		f_{\!_{B4}}^{y} &
		m_{\!_{B4}}^{z} &
		f_{\!_{B3}}^{x} &
		f_{\!_{B3}}^{y}
	\end{bmatrix}^{\top} \\
	\hat{\F}_{\!\psi} & \hspace*{-0.1cm}=\hspace*{-0.1cm} & \begin{bmatrix}
		\widetilde{m}_{\!_{B1}}^{z} \!\!&\!\!
		-\hat{f}_{\!_{B4}}^{x} \!\!&\!\!
		-\hat{f}_{\!_{B4}}^{y} \!\!&\!\!
		-\hat{m}_{\!_{B4}}^{z} \!\!&\!\!
		-\hat{m}_{\!_{B3}}^{z} \!&\!
		\hat{f}_{\!_{B3}}^{x} \!&\!
		\hat{f}_{\!_{B3}}^{y}
	\end{bmatrix}^{\top}
\end{eqnarray}
{\footnotesize \begin{equation}
		\bld{A} \ = \ \begin{bmatrix}
			-L_{1}\sin \!q_{2} \!&\! L_{1}\cos \!q_{2} \!&\! 0 \!&\! 0 \!&\! 0 \!&\! 0 \!&\! 0 \\
			1 \!&\! 0 \!&\! -1 \!&\! 0 \!&\! 0 \!&\! 0 \!&\! 0 \\
			0 \!&\! 1 \!&\! 0 \!&\! -1 \!&\! 0 \!&\! 0 \!&\! 0 \\
			0 \!&\! L_c \!&\! 0 \!&\! 0 \!&\! -1 \!&\! 0 \!&\! 0 \\
			0 \!&\! 0 \!&\! 0 \!&\! x\!+\!L_{c0} \!&\! 1 \!&\! 0 \!&\! 0 \\
			0 \!&\! 0 \!&\! 1 \!&\! 0 \!&\! 0 \!&\! -1 \!&\! 0 \\
			0 \!&\! 0 \!&\! 0 \!&\! 1 \!&\! 0 \!&\! 0 \!&\! -1 \\
		\end{bmatrix}
\end{equation}}
where $\widetilde{\F}_{\!_{\!B1}} = \hat{\F}_{\!_{\!B1}} + \mbox{Ad}_{G_{\!_{E}}^{\!_{B1}}}^{*} \F_{\!_{\!E}}$ and $\widetilde{m}_{\!_{B1}}^{z} = \bld{s}_{z}^{\top}\widetilde{\F}_{\!_{\!B1}}$.

Therefore, the inverse dynamics can be solved by the Cramer rule as
\begin{equation}
	f_{\!_{B4}}^{x} \ = \ \det \bld{A}_{\psi} / \det \bld{A}
	\label{eq:id}
\end{equation}
where $\bld{A}_{\psi}$ is $\bld{A}$ but with its third column equal to $\hat{\F}_{\!\psi}$. Then, this solution exists when
\begin{equation*}
	\det \bld{A} \ = \ -L_{1}\sin \!q_{2} \left( x + x_0 \right) \ \neq \ 0
\end{equation*}
which is ensured, since $L_{1} > 0$, $q_2 \neq0$, and $x > 0$.
\section{Total wrench in the linear-actuator reference frame}
\label{app:force_piston}
Assumption (\ref{eq:assum_02}) establishes that wrenches have zero elements from their third to their fifth entries. This can justify an equivalent projection $se^{*}(3)\rightarrow se^{*}(2)$ that facilitates the local analysis of the parallel mechanism by neglecting the zero elements. We consider that all wrenches in (\ref{eq:assum_02}) can be projected to a lower-dimensional dual Lie algebra by means of the projector $\bld{\Xi}:se^{*}(3)\rightarrow se^{*}(2)$
\begin{equation*}
	\bld{\Xi} \ = \ {\footnotesize \begin{bmatrix}
			1 & 0 & 0 & 0 & 0 & 0 \\
			0 & 1 & 0 & 0 & 0 & 0 \\
			0 & 0 & 0 & 0 & 0 & 1
	\end{bmatrix}}
\end{equation*}
and $\bld{\Xi}^{\top}:se^{*}(2)\rightarrow se^{*}(3)$.

Equation (\ref{eq:id}) solves for the actuator force in $x$ axis. Thus, we can compute the three non zero elements of $\F_{\!_{\!B4}}$, see (\ref{eq:56}) by using the linear system (\ref{eq:linSys}), as
\begin{equation*}
	\begin{bmatrix}
		f_{\!_{B4}}^{x} \\
		f_{\!_{B4}}^{y} \\
		m_{\!_{B4}}^{z}
	\end{bmatrix} \ = \ \bld{A}_{[3:5,:]}^{-1}\hat{\F}_{\!\psi} \\ \ = \ \bld{A}_{[3:5,:]}^{-1}\bld{C}\begin{bmatrix}
		\bld{\Xi}\widetilde{\F}_{\!_{\!B1}} \\
		\bld{\Xi}\hat{\F}_{\!_{\!B3}} \\
		\bld{\Xi}\hat{\F}_{\!_{\!B4}}
	\end{bmatrix} \\
\end{equation*}
where $\bld{A}_{[3:5,:]}^{-1}\in\real^{3\times7}$ stands for three rows, according to M{\footnotesize ATLAB} notation, of the inverse matrix of \bld{A} and $\bld{C}\in\real^{7\times 9}$ is a constant commutation matrix that rearranges the elements in $\hat{\F}_{\!\psi}$.

The block matrix $\bld{A}_{[3:5,:]}^{-1}\bld{C} \in \real^{3\times 9}$ is composed by three blocks $\bar{\bld{K}}_{i}\in\real^{3\times3}$ as follows
\begin{eqnarray}
	\bld{A}_{[3:5,:]}^{-1}\bld{C} & \hspace*{-0.2cm}=\hspace*{-0.2cm} & \begin{bmatrix}
		\bar{\bld{K}}_{1} & \bar{\bld{K}}_{3} & \bar{\bld{K}}_{4}
	\end{bmatrix} \label{eq:39} \\
	& \hspace*{-0.2cm}=\hspace*{-0.2cm} & {\footnotesize \begin{bmatrix}
			0 \!\!&\!\! 0 \!\!&\!\! k_b^{-1} \!\!&\!\! 0 \!\!&\!\! 0 \!\!&\!\! k_f \!\!&\!\! 1 \!\!&\!\! k_a k_f \!\!&\!\! k_f \\
			0 \!\!&\!\! 0 \!\!&\!\! 0 \!\!&\!\! 0 \!\!&\!\! 0 \!\!&\!\! -k_e^{-1} \!\!&\!\! 0 \!\!&\!\! k_g \!\!&\!\! k_e^{-1} \\
			0 \!\!&\!\! 0 \!\!&\!\! 0 \!\!&\!\! 0 \!\!&\!\! 0 \!\!&\!\! - k_g \!\!&\!\! 0 \!\!&\!\! -k_a k_g \!\!&\!\! k_a k_e^{-1}
	\end{bmatrix}}
\end{eqnarray}
where $k_a = x\!+L_{c0}$, $k_b = -L_{1}\!\sin \!q_{2}$, $k_c = L_{1}\!\cos \!q_{2}$, $k_d = k_b(k_a+L_{c})$, $k_e = k_a + L_{c}$, $k_f = k_c k_d^{-1}$ and $k_g = L_{c}k_e^{-1}$.

If we project the matrices $\bar{\bld{K}}_{i}$ from the Lie algebra $se^{*}(2)$ to $se^{*}(3)$ we get $\bld{K}_{i}\in\real^{6\times6}$ as
\begin{equation}
	\bld{K}_{i} \ = \ \bld{\Xi}^{\top} \bar{\bld{K}}_{i} \ \bld{\Xi}
	\label{eq:40}
\end{equation}
and we can write the total actuator wrench $\F_{\!_{\!B4}}$ as a combination of the wrenches $\widetilde{\F}_{\!_{\!B1}}$, $\hat{\F}_{\!_{\!B3}}$, and $\hat{\F}_{\!_{\!B4}}$ by the expression
\begin{equation}
	\F_{\!\!_{\!B4}} \ = \ \bld{K}_{\!1} \widetilde{\F}_{\!\!_{\!B1}} + \bld{K}_{\!3} \hat{\F}_{\!\!_{\!B3}} + \bld{K}_{\!4} \hat{\F}_{\!\!_{\!B4}}
	\label{eq:99}
\end{equation}
%


\bibliographystyle{elsarticle-num} 
\bibliography{biblio.bib}


%
%
%
\end{document}